\documentclass{arxiv_article}

\usepackage{amsmath,amssymb,amsthm}
\usepackage{xspace}
\usepackage{fancyvrb}
\usepackage{array}
\usepackage{tabularx}
\usepackage{ragged2e}

\usepackage{booktabs}
\usepackage{colortbl}
\usepackage{mathtools}
\usepackage{xcolor}
\usepackage{algorithm}
\usepackage{algpseudocode}

\usepackage{subcaption}
\usepackage{adjustbox}
\usepackage{nicematrix}
\usepackage{makecell}
\usepackage{graphicx}

\theoremstyle{plain}
\newtheorem{theorem}{Theorem}[]

\newtheorem{assumption}[theorem]{Assumption}

\usepackage{wrapfig}
\usepackage{float}

\newcommand{\REQUIRE}{\Require}
\newcommand{\STATE}{\State}
\newcommand{\FOR}[1]{\For{#1}}
\newcommand{\ENDFOR}{\EndFor}
\newcommand{\WHILE}[1]{\While{#1}}
\newcommand{\ENDWHILE}{\EndWhile}
\newcommand{\IF}[1]{\If{#1}}
\newcommand{\ENDIF}{\EndIf}

\definecolor{lightpink}{rgb}{1,0.9,0.9}
\definecolor{lightblue}{rgb}{0.8,0.9,1.0}

\newcolumntype{C}[1]{>{\centering\arraybackslash}p{#1}}
\definecolor{oatpink}{HTML}{FDE2E4}
\definecolor{mypurple}{HTML}{4A148C}
\usepackage[svgnames]{xcolor}
\usepackage[most]{tcolorbox}

\usepackage[textsize=tiny]{todonotes}
\usepackage{calc}
\usepackage[table]{xcolor}
\usepackage[dvipsnames]{xcolor}
\definecolor{highlightred}{HTML}{FFF0F0}
\definecolor{highlightblue}{RGB}{236,244,252}
\definecolor{caseStudyGreenDark}{HTML}{2E8B57} 
\definecolor{caseStudyGreenLight}{HTML}{F0FFF0} 
\definecolor{answerRed}{HTML}{DC143C}         
\definecolor{answerBlue}{HTML}{0000CD}        
\definecolor{confidenceGray}{HTML}{8B4513} 

\definecolor{cGreen}{HTML}{336600}
\definecolor{cgray}{HTML}{FAFAFA}
\definecolor{graybg}{gray}{0.95}

\renewcommand\affiliationformat[2][]{ {\footnotesize $^{#1}$#2} }
\title{\LARGE ReGuLaR: Variational Latent Reasoning Guided by Rendered Chain-of-Thought}
\author[1,2]{Fanmeng Wang}
\author[1]{Haotian Liu}
\author[2]{Guojiang Zhao}
\author[1,3,4]{Hongteng Xu\thanks{Corresponding authors. Email: hongtengxu@ruc.edu.cn and gaozf@dp.tech}}
\author[2]{Zhifeng Gao\samethanks[1]}
\affiliation[1]{Gaoling School of Artificial Intelligence, Renmin University of China}
\affiliation[2]{DP Technology}
\affiliation[3]{Beijing Key Laboratory of Research on Large Models and Intelligent Governance}
\affiliation[4]{Engineering Research Center of Next-Generation Intelligent Search and Recommendation, MOE}

\begin{document}
\abstract{While Chain-of-Thought (CoT) significantly enhances the performance of Large Language Models (LLMs), explicit reasoning chains introduce substantial computational redundancy. 
Recent latent reasoning methods attempt to mitigate this by compressing reasoning processes into latent space, but often suffer from severe performance degradation due to the lack of appropriate compression guidance.
In this study, we propose \textbf{Re}ndered CoT-\textbf{Gu}ided variational \textbf{La}tent \textbf{R}easoning (\textbf{ReGuLaR}), a simple yet novel latent learning paradigm resolving this issue.
Fundamentally, we formulate latent reasoning within the Variational Auto-Encoding (VAE) framework, sampling the current latent reasoning state from the posterior distribution conditioned on previous ones.
Specifically, when learning this variational latent reasoning model, we render explicit reasoning chains as images, from which we extract dense visual-semantic representations to regularize the posterior distribution, thereby achieving efficient compression with minimal information loss.
Extensive experiments demonstrate that ReGuLaR significantly outperforms existing latent reasoning methods across both computational efficiency and reasoning effectiveness, and even surpasses CoT through multi-modal reasoning, providing a new and insightful solution to latent reasoning.
Code: \hypersetup{urlcolor=mypurple}\href{https://github.com/FanmengWang/ReGuLaR}{https://github.com/FanmengWang/ReGuLaR}.}

\maketitle

\section{Introduction}
Large Language Models (LLMs) have demonstrated exceptional performance in solving complex problems, a success largely attributed to the adoption of Chain-of-Thought (CoT) techniques~\citep{wei2022chain, jin2024impact_cot2}.
By eliciting LLMs to generate intermediate reasoning steps in natural language, CoT effectively decomposes complex problems, significantly bolstering accuracy on challenging queries~\citep{fei2023reasoning, wang2024leveraging}. 
However, such reasoning processes suffer from inherent inefficiency because they rely on explicit token-by-token generation, and many tokens can be redundant for improving reasoning~\citep{kang2025c3ot, sui2025stop}.
This results in prohibitive computational overhead and increased inference latency, fundamentally limiting the scalability of LLM reasoning.

In this context, recent studies have explored latent reasoning as a compelling alternative to explicit CoT~\citep{zhu2025survey}.
By operating directly on continuous representations, latent reasoning compresses reasoning processes into the high-dimensional latent space, thereby circumventing the overhead of decoding intermediate reasoning tokens~\citep{zhu2025reasoning}. 
To instantiate this paradigm, several representative frameworks have been proposed, e.g., Coconut~\citep{coconut_hao2024training} and CoLaR~\citep{colar_tan2025think}.
However, while alleviating computational burdens, existing latent reasoning methods often suffer from severe performance degradation, primarily because \textit{the compression of reasoning processes lacks appropriate guidance.} 
Specifically, these methods typically rely on recursively or dynamically utilizing the hidden states of reasoning tokens to propagate logical dependencies. 
In the absence of discrete tokens to anchor the reasoning trajectory, this unconstrained recursive process becomes highly susceptible to error accumulation, leading to significant information loss and semantic drift.

\begin{wrapfigure}{r}{0.5\textwidth}
    \centering
    \vspace{-10pt} 
    \includegraphics[width=\linewidth]{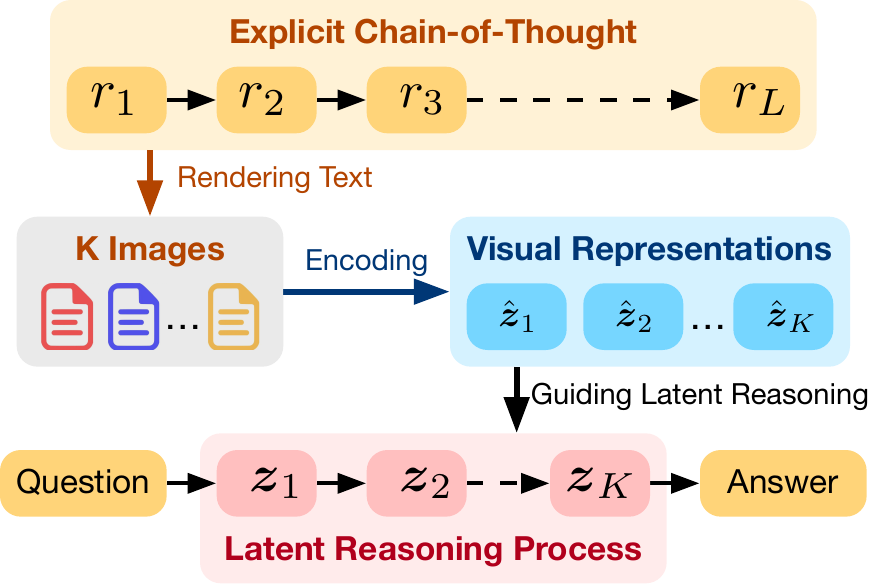}
    \caption{Illustration of our modeling principle. 
    Given an explicit reasoning chain of length $L$, we render it onto $K$ images ($K \ll L$) and extract corresponding visual representations to guide the latent reasoning process with $K$ steps.}
    \label{fig: Concept}
    \vspace{-10pt} 
\end{wrapfigure}

In this study, we propose a novel and insightful paradigm to resolve the above challenge, learning a \textbf{Re}ndered CoT-\textbf{Gu}ided variational \textbf{La}tent \textbf{R}easoning (\textbf{ReGuLaR}) model. 
Fundamentally, we formulate latent reasoning as a probabilistic modeling task within the Variational Auto-Encoding (VAE) framework~\citep{kingma2014auto}.
In this formulation, the latent reasoning process is achieved by sampling the current latent reasoning state from the posterior distribution conditioned on previous ones.
Here, we optimize this model by maximizing its Evidence Lower Bound (ELBO)~\citep{neal1998view}, wherein the prior distribution of the latent reasoning state plays a critical role in regularizing the posterior distribution.
As illustrated in Figure~\ref{fig: Concept}, we render the explicit reasoning chain as images and then leverage the visual encoder to extract visual representations with dense semantics.
This rendering step is lossless, so we utilize these visual representations to regularize the posterior distribution of the latent reasoning state during training, thereby leading to our ReGuLaR with compressed but semantically meaningful latent reasoning states. 

To the best of our knowledge, ReGuLaR is the first work that applies the VAE framework to understanding and modeling latent reasoning.
With this framework, we demonstrate the importance of the latent reasoning state prior and propose a promising approach to designing a semantically meaningful, information-preserving prior for latent reasoning states. 
Extensive experiments demonstrate that ReGuLaR provides a new and insightful solution to latent reasoning. 
Specifically, it significantly outperforms existing latent reasoning methods, achieving state-of-the-art performance with minimal reasoning length. 
Furthermore, ReGuLaR natively supports multi-modality within its latent reasoning processes by rendering non-textual elements alongside text, enabling it to surpass explicit CoT in complicated reasoning scenarios.

\section{Related Work}
\subsection{LLM Reasoning and Latent Reasoning}
The reasoning capabilities of LLMs have been advanced by CoT techniques, which prompt the generation of intermediate reasoning steps in natural language~\citep{wei2022chain, shao2024deepseekmath, jin2024impact_cot2}. 
Building on this, subsequent research has explored various CoT constructions, including Tab-CoT~\citep{ziqi2023tab}, ToT~\citep{yao2023tree}, and GoT-Rationale~\citep{besta2024graph}. 
While these methods manifest reasoning in different explicit forms, verbose intermediate reasoning steps inevitably incur substantial computational cost and inference latency~\citep{kang2025c3ot, sui2025stop}.

To mitigate this bottleneck, iCoT~\citep{deng2024explicit} internalizes intermediate reasoning steps by progressively removing them during training.
Moreover, the latent reasoning paradigm has emerged, which transforms intermediate reasoning tokens into continuous representations and eliminates the overhead of language-decoding steps by executing latent reasoning processes~\citep{zhu2025survey}. 
In particular, Coconut~\citep{coconut_hao2024training} pioneers this direction by recursively utilizing the last-layer hidden states of LLMs as the continuous latent thought, functioning as the next input embedding to drive subsequent reasoning.
Meanwhile, CODI~\citep{shen2025codi} further employs self-distillation to align the hidden activations of latent thoughts with explicit CoT trajectories.
Most recently, CoLaR~\citep{colar_tan2025think} achieves state-of-the-art performance by leveraging training mechanisms with variable compression factors to support flexible reasoning length.
However, these methods suffer from severe performance degradation compared with explicit CoT, primarily due to the lack of appropriate compression guidance, thereby limiting their practical utility.

\subsection{Empowering LLMs via Visual-Text Compression}
LLMs typically rely on discrete tokenization to process text inputs, a mechanism that inevitably fragments the global semantic topology while incurring substantial computational cost~\citep{zhao2023survey}.
To overcome these inefficiencies, the paradigm of visual-text compression~\citep{zhao2025vtcbench} has been explored.
It renders textual content as images and embeds them via the visual encoder, thereby exploiting the high information density of the visual modality.
In particular, VisInContext~\citep{wang2024leveraging} introduces a visualized in-context text processing framework that leverages compact visual tokens to replace long textual contexts, effectively expanding context windows without additional computational burden.
Subsequently, VIST~\citep{xing2025vision_vist} proposes a fast-path compression mechanism that leverages the lightweight visual encoder to process rendered images of distant contexts for rapid skimming, significantly improving efficiency.
Recently, DeepSeek-OCR~\citep{wei2025deepseek_OCR} further brings this paradigm to the forefront by validating its feasibility and scalability on massive textual data, enabling the mapping of extensive textual contexts into ultra-compact visual tokens with high compression ratios.

While these works primarily focus on compressing input contexts, they provide strong evidence for visual representations as high-density carriers of textual information.
Therefore, such visual-text compression should also be useful in empowering latent reasoning, as our work verifies.

\section{Proposed Method}
\subsection{Problem Statement and Preliminaries}\label{ssec: problem}
Formally, suppose that we have a reasoning dataset $\mathcal{D} = \{(\mathcal{Q}, \mathcal{R}, \mathcal{A})\}$, in which each tuple contains an input question $\mathcal{Q}$, an intermediate reasoning chain $\mathcal{R}$, and the final answer $\mathcal{A}$.
Here, we represent each element in the tuple as a token sequence, i.e., $\mathcal{Q}=\{q_{i}\}_{i=1}^{L_q}$, $\mathcal{R}=\{r_{i}\}_{i=1}^{L_r}$, and $\mathcal{A}=\{a_{i}\}_{i=1}^{L_a}$, where $L_q$, $L_r$, and $L_a$ are the respective sequence lengths.
Additionally, for an arbitrary token sequence $\mathcal{T}$, we denote the corresponding subsequence before the $i$-th token as $\mathcal{T}_{<i}$.

\textbf{Learning an LLM with explicit CoT.}
Under the Chain-of-Thought (CoT) paradigm, the LLM explicitly generates the reasoning chain token by token before producing the final answer, thereby bridging the logical gap between the question and the corresponding answer.
Accordingly, given $(\mathcal{Q},\mathcal{R},\mathcal{A})$, we can learn the LLM via the Maximum Likelihood Estimation (MLE) as follows: 
\begin{eqnarray}\label{eq: cot}
    \max_{\theta}\underbrace{\sum_{i=1}^{L_r}\log p_{\theta}(r_i|\mathcal{Q},\mathcal{R}_{<i})}_{\mathcal{L}_\text{reasoning}}\!+\!\underbrace{\sum_{i=1}^{L_a}\log p_{\theta}(a_i|\mathcal{Q},\mathcal{R},\mathcal{A}_{<i})}_{\mathcal{L}_{\text{answer}}},\!\!\!\!\!
\end{eqnarray}
where $\theta$ denotes the model parameters, $p_{\theta}$ represents the conditional probability of each token given its history. 
In practice, we implement $\mathcal{L}_\text{reasoning}$ and $\mathcal{L}_\text{answer}$ as two Cross-Entropy (CE) losses~\citep{mao2023cross}.

\begin{wrapfigure}{r}{0.5\textwidth}
    \centering
    \vspace{-2pt} 
    \begin{subfigure}{0.49\linewidth} 
        \centering
        \includegraphics[width=\linewidth]{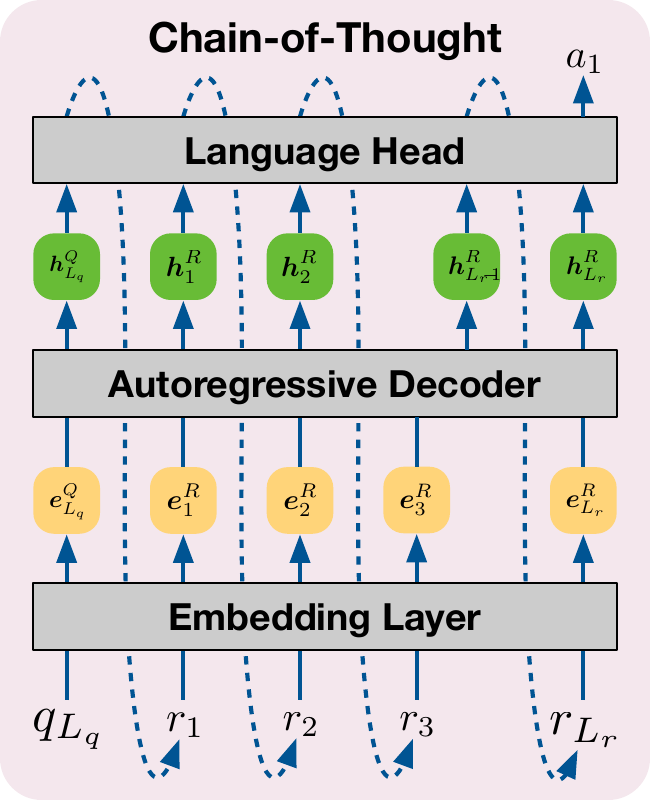} 
        \caption{Explicit reasoning}
        \label{fig:paradigm_a}
    \end{subfigure}
    \hfill 
    \begin{subfigure}{0.49\linewidth}
        \centering
        \includegraphics[width=\linewidth]{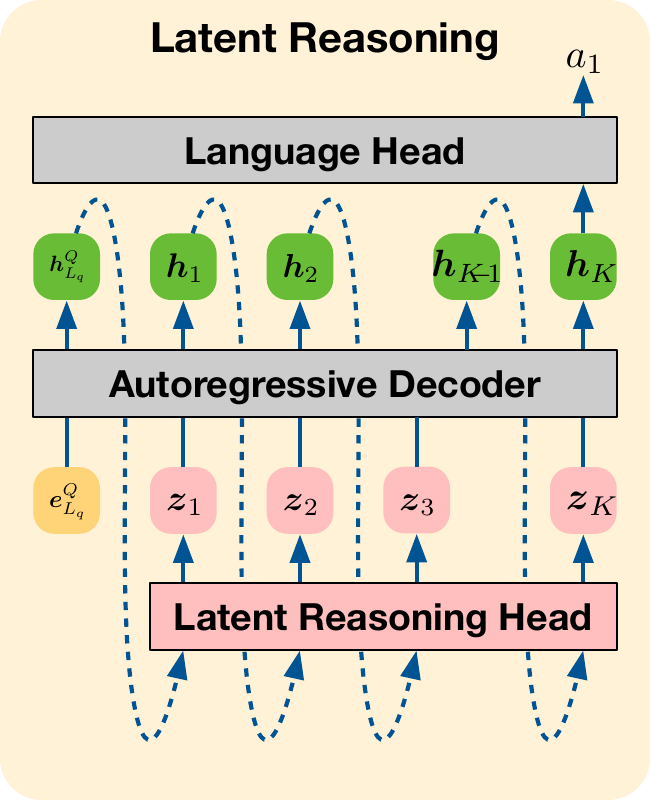} 
        \caption{Latent reasoning}
        \label{fig:paradigm_b}
    \end{subfigure}
    \vspace{-2pt} 
    \caption{Comparison of CoT-based explicit reasoning and latent reasoning, where the autoregressive decoder corresponds to the underlying LLM.}
    \label{fig: Paradigm}
\end{wrapfigure}

As illustrated in Figure~\ref{fig:paradigm_a}, for the LLM using explicit reasoning, all tokens are first mapped into continuous embedding vectors (denoted as $\bm{e}^{Q}_{1:L_q}$, $\bm{e}^{R}_{1:L_r}$, and $\bm{e}^{A}_{1:L_a}$), and subsequently transformed by the model layers into the last hidden states (denoted as $\bm{h}^{Q}_{1:L_q}$, $\bm{h}^{R}_{1:L_r}$, and $\bm{h}^{A}_{1:L_a}$).
In this context, the LLM necessitates the explicit token-by-token generation of the intermediate reasoning chain during inference, resulting in substantial computational overhead and inference latency.

\textbf{Learning an LLM with latent reasoning.}
The inefficiency of CoT further motivates the latent reasoning paradigm illustrated in Figure~\ref{fig:paradigm_b}. 
Specifically, latent reasoning replaces discrete reasoning tokens $\mathcal{R}=\{r_{i}\}_{i=1}^{L_r}$ with continuous latent reasoning states $\bm{Z}=\{\bm{z}_{k}\}_{k=1}^{K}$.
In this context, the LLM employs an additional latent reasoning head to derive the latent reasoning state from the current hidden state, which functions as the subsequent input embedding, thereby eliminating the overhead of decoding those intermediate reasoning tokens. 
Moreover, the sequence of latent reasoning states can be much shorter than the corresponding reasoning chain ($K \ll L_r$), which helps improve inference efficiency significantly.

Learning such a latent reasoning model corresponds to the following optimization problem, i.e., 
\begin{equation}\label{eq:latent_answer_loss} 
\sideset{}{_{\theta'=\{\theta,\tau\}}}\max\sideset{}{_{i=1}^{L_a}}\sum \log p_{\theta'}(a_i \mid \mathcal{Q}, \bm{Z}, \mathcal{A}_{<i}), 
\end{equation} 
where the final answer is conditioned on the latent reasoning state sequence $\bm{Z}$ rather than the explicit reasoning chain $\mathcal{R}$. 
The parameter $\theta'=\{\theta,\tau\}$, where $\tau$ denotes the parameters of the latent reasoning head.

Compared to~\eqref{eq: cot}, this optimization problem is inherently challenging due to the absence of ground-truth supervision for latent reasoning states. 
To mitigate this, Coconut~\citep{coconut_hao2024training} employs the multi-stage curriculum to progressively replace discrete reasoning steps with the last hidden state of the preceding context. 
However, since it relies on distilling knowledge from the original CoT, its performance is fundamentally bounded.
Furthermore, CoLaR~\citep{colar_tan2025think} directly constructs latent reasoning states by dynamically compressing the embeddings of original reasoning tokens.
Nevertheless, its token grouping strategy introduces arbitrary inductive biases, and this simple aggregation inevitably leads to semantic information loss.

To overcome these problems and achieve effective latent reasoning, \textbf{we need to regularize latent reasoning states, and further impose additional information during training to ensure semantically meaningful}, which motivates the proposed ReGuLaR method.

\subsection{The Variational Latent Reasoning Framework}\label{ssec: vae}
Suppose that we would like to learn an LLM with the latent reasoning mechanism, employing the latent reasoning state sequence $\bm{Z}$ of length $K$ to replace the explicit reasoning chain $\mathcal{R}$ of length $L_r$. 
For such an LLM, we decompose its parameters into two parts, i.e., $\theta'=\{\psi,\phi\}$, where $\psi$ denotes the parameters of the language head that outputs discrete tokens and $\phi$ denotes the remaining parameters that comprise the latent reasoning head deriving $\bm{Z}$. 
In this context, we build a \textit{variational latent reasoning process}: the LLM samples each latent reasoning state from its posterior distribution given the question and the previous ones, i.e., 
\begin{eqnarray}\label{eq:posterior}
\begin{aligned}
    \bm{z}_k\sim p_{\phi}(\cdot|\mathcal{Q},\bm{Z}_{<k}),~\text{for}~k=1,...,K,
\end{aligned}
\end{eqnarray}
where $\bm{Z}_{<1}=\emptyset$.
In this study, we model $p_{\phi}(\cdot|\mathcal{Q},\bm{Z}_{<k})$ as a normal distribution $\mathcal{N}(\bm{\mu}_k,\text{diag}(\bm{\sigma}_k^2))$.
Applying the reparametrization trick~\citep{kingma2014auto}, we leverage the latent reasoning head of the LLM to output $\bm{\mu}_k$ and $\log\bm{\sigma}_k$ based on $\mathcal{Q}$ and $\bm{Z}_{<k}$, and then sample $\bm{z}_k=\bm{\mu}_k+\bm{\sigma}_k\odot\bm{\epsilon}$, where $\bm{\epsilon}\sim\mathcal{N}(\bm{0},\bm{I})$ and $\odot$ denotes the Hadamard product operation.

\textbf{Desideratum.}
Ideally, $\bm{Z}$ should have the same information as the original reasoning chain $\mathcal{R}$.
More specifically, for $k=1,...,K$, the latent reasoning state $\bm{z}_k$ should correspond to the $k$-th segment of $\mathcal{R}$, denoted as $\mathcal{R}_k$, where $\mathcal{R}_k\cap\mathcal{R}_{k'}=\emptyset$ for $k\neq k'$ and $\mathcal{R}=\cup_{k=1}^{K}\mathcal{R}_k$. 
Accordingly, we should be able to sample tokens in $\mathcal{R}_k$ through the distribution conditioned on $\bm{z}_k$, i.e., 
\begin{eqnarray}\label{eq:decoding}
\begin{aligned}
    r\sim p_{\psi}(\cdot|\bm{z}_{k})~\text{and}~r\in\mathcal{R}_k,~\text{for}~k=1,...,K.
\end{aligned}   
\end{eqnarray}
Here, $p_{\psi}$ is modeled using Softmax Regression, and the language head of the LLM is used to output corresponding logit values of sampled tokens.

Motivated by the above desideratum, we introduce the following conditional independence assumption.
\begin{assumption}[Conditional Independence]\label{assume:independence}
For $k=1,...,K$, $i)$ the token $r\in\mathcal{R}_k$ is independent with tokens in $\{\mathcal{Q},\mathcal{R}\}\setminus\mathcal{R}_k$ conditioned on the latent reasoning state $\bm{z}_k$, and $ii)$ the latent reasoning state $\bm{z}_k$ is independent with tokens in $\{\mathcal{Q},\mathcal{R}\}\setminus\mathcal{R}_k$ conditioned on the tokens in $\mathcal{R}_k$.
\end{assumption}

Suppose that $r$ is the $i$-th token in $\mathcal{R}$, which corresponds to the $j$-th token in $\mathcal{R}_k$. 
For the conditional probability used in explicit CoT, we can rewrite it based on the assumption:
\begin{eqnarray}\label{eq:decomposition_r}
\begin{aligned}
    p(r|\mathcal{Q},\mathcal{R}_{<i})
    =&\int_{\bm{z}_k}p(r|\bm{z}_k,\mathcal{Q},\mathcal{R}_{<i})p(\bm{z}_k|\mathcal{Q},\mathcal{R}_{<i})\mathrm{d}\bm{z}_k\!\!\\
    =&\int_{\bm{z}_k}p_{\psi}(r|\bm{z}_k)p_{\gamma}(\bm{z}_k|\mathcal{R}_{k,<j})\mathrm{d}\bm{z}_k.
\end{aligned}    
\end{eqnarray}
Here, $p_{\psi}(r|\bm{z}_k)$ is the probability of the token conditioned on $\bm{z}_k$, which is parametrized by the language head of the LLM.
$p_{\gamma}(\bm{z}_k|\mathcal{R}_{k,<j})$ is the distribution of $\bm{z}_k$ conditioned on the partial information (i.e., before the $j$-th token) of the reasoning segment $\mathcal{R}_{k}$, whose parameters are denoted as $\gamma$. 

Furthermore, when considering the posterior distribution in~\eqref{eq:posterior}, we can derive the Evidence Lower Bound (ELBO) for $\log p(r|\mathcal{Q},\mathcal{R}_{<i})$ as follows:
\begin{eqnarray}\label{eq:elbo}
\begin{aligned}
    \log p(r|\mathcal{Q},\mathcal{R}_{<i}) 
    \geq\, &\mathbb{E}_{\bm{z}_k\sim p_{\phi}(\cdot|\mathcal{Q},\bm{Z}_{<k})}\Big[\log\frac{p_{\psi}(r|\bm{z}_k)p_{\gamma}(\bm{z}_k|\mathcal{R}_{k,<j})}{p_{\phi}(\bm{z}_k|\mathcal{Q},\bm{Z}_{<k})}\Big] \\
    =\, &\mathbb{E}_{\bm{z}_k\sim p_{\phi}(\cdot|\mathcal{Q},\bm{Z}_{<k})}[\log p_{\psi}(r|\bm{z}_k)] 
    -\text{KL}[\underbrace{p_{\phi}(\cdot|\mathcal{Q},\bm{Z}_{<k})}_{\text{Posterior of $\bm{z}_k$}}\,\|\, \underbrace{p(\cdot|\mathcal{R}_{k,<j})}_{\text{Prior of $\bm{z}_k$}}].
\end{aligned}
\end{eqnarray}

The ELBO of $\log p(r|\mathcal{Q},\mathcal{R}_{<i})$ naturally leads to a variational auto-encoding framework of latent reasoning. 
For the LLM, its autoregressive module with the latent reasoning head, whose parameters are $\phi$, works as the encoder embedding the question and the previous latent reasoning states to the current latent reasoning state.
Its language head $\psi$ works as the decoder generating tokens based on the latent reasoning states.
In~\eqref{eq:elbo}, the first term corresponds to the latent reasoning loss, measuring the likelihood of reasoning tokens given the sampled latent reasoning states.
The second term is the KL divergence between the posterior and prior distributions, regularizing the posterior distribution.

Considering the ELBO in~\eqref{eq:elbo} with the loss in~\eqref{eq:latent_answer_loss}, we can learn a variational latent reasoning model by solving the following optimization problem:
\begin{eqnarray}\label{eq: learning}
\begin{aligned}
    \max_{\theta'=\{\phi,\psi\},\gamma}&\sideset{}{_{i=1}^{L_a}}\sum\underbrace{\mathbb{E}_{\bm{Z}\sim p_{\phi}(\cdot|\mathcal{Q})}[\log p_{\theta'}(a_i | \mathcal{Q}, \bm{Z}, \mathcal{A}_{<i})]}_{\mathcal{L}_\text{answer}^{\text{Latent}}}\\
    &+\sideset{}{_{k=1}^{K}}\sum\sideset{}{_{r_j\in\mathcal{R}_k}}\sum\underbrace{\mathbb{E}_{\bm{z}_k\sim p_{\phi}(\cdot|\mathcal{Q},\bm{Z}_{<k})}[\log p_{\psi}(r_j|\bm{z}_k)]}_{\mathcal{L}_\text{reasoning}^{\text{Latent}}}\!\!\\
    &-\sideset{}{_{k=1}^{K}}\sum\underbrace{\text{KL}[p_{\phi}(\cdot|\mathcal{Q},\bm{Z}_{<k})\|p_{\gamma}(\cdot|\mathcal{R}_{k})]}_{\text{Regularizer of posterior}}.
\end{aligned}
\end{eqnarray}
Here, $\bm{Z}\sim p_{\phi}(\cdot|\mathcal{Q})$ is implemented by sequential sampling shown in~\eqref{eq:posterior}.
$p_{\gamma}(\cdot|\mathcal{R}_k)$ denotes the distribution of $\bm{z}_k$ conditioned on $\mathcal{R}_k$, whose parameters are denoted as $\gamma$.
Similar to~\citep{dilokthanakul2016deep,tomczak2018vae,xu2020learning}, we learn the prior distribution of the latent reasoning state $p_{\gamma}$ together with the latent reasoning model.
Obviously, this learning problem is analogous to that of explicit CoT in~\eqref{eq: cot}, which maximizes the likelihoods of both reasoning and answer tokens. 
Furthermore, the posterior distribution $p_{\phi}$ is optimized under the guidance of $p_{\gamma}$.

\textbf{Remark.} In~\eqref{eq: learning}, we replace the $p(\cdot|\mathcal{R}_{k,<j})$ in~\eqref{eq:elbo} with $p_{\gamma}(\cdot|\mathcal{R}_k)$.
Such a modification leads to a ``stable'' prior distribution invariant with the selection of the reasoning token $r_j\in\mathcal{R}_k$, which is reasonable in practice.
Firstly, as aforementioned, an ideal latent reasoning state $\bm{z}_k$ should cover the information of $\mathcal{R}_k$, so that modeling its distribution conditioned on $\mathcal{R}_k$ rather than $\mathcal{R}_{k,<j}$ can impose more information to the model, leading to better regularization.
Secondly, if the distribution $p_{\gamma}$ changes with respect to the selection of the reasoning tokens, we would have to recompute the KL divergence in~\eqref{eq: learning} for each $r_j\in\mathcal{R}_k$, which would cause significant training overhead.
Therefore, applying $p_{\gamma}(\cdot|\mathcal{R}_k)$ is reasonable and  efficient in practice. 

\subsection{Implementing the Framework via ReGuLaR}\label{ssec: implement}
As analyzed in Section~\ref{ssec: problem}, the crux of learning latent reasoning models lies in guiding the posterior of latent reasoning states with less information loss.
Therefore, designing and learning the semantically meaningful prior distribution $p_{\gamma}$ is critical for our variational latent reasoning model.

\begin{wrapfigure}{r}{0.46\textwidth}
    \centering
    \vspace{-10pt} 
    \includegraphics[width=\linewidth]{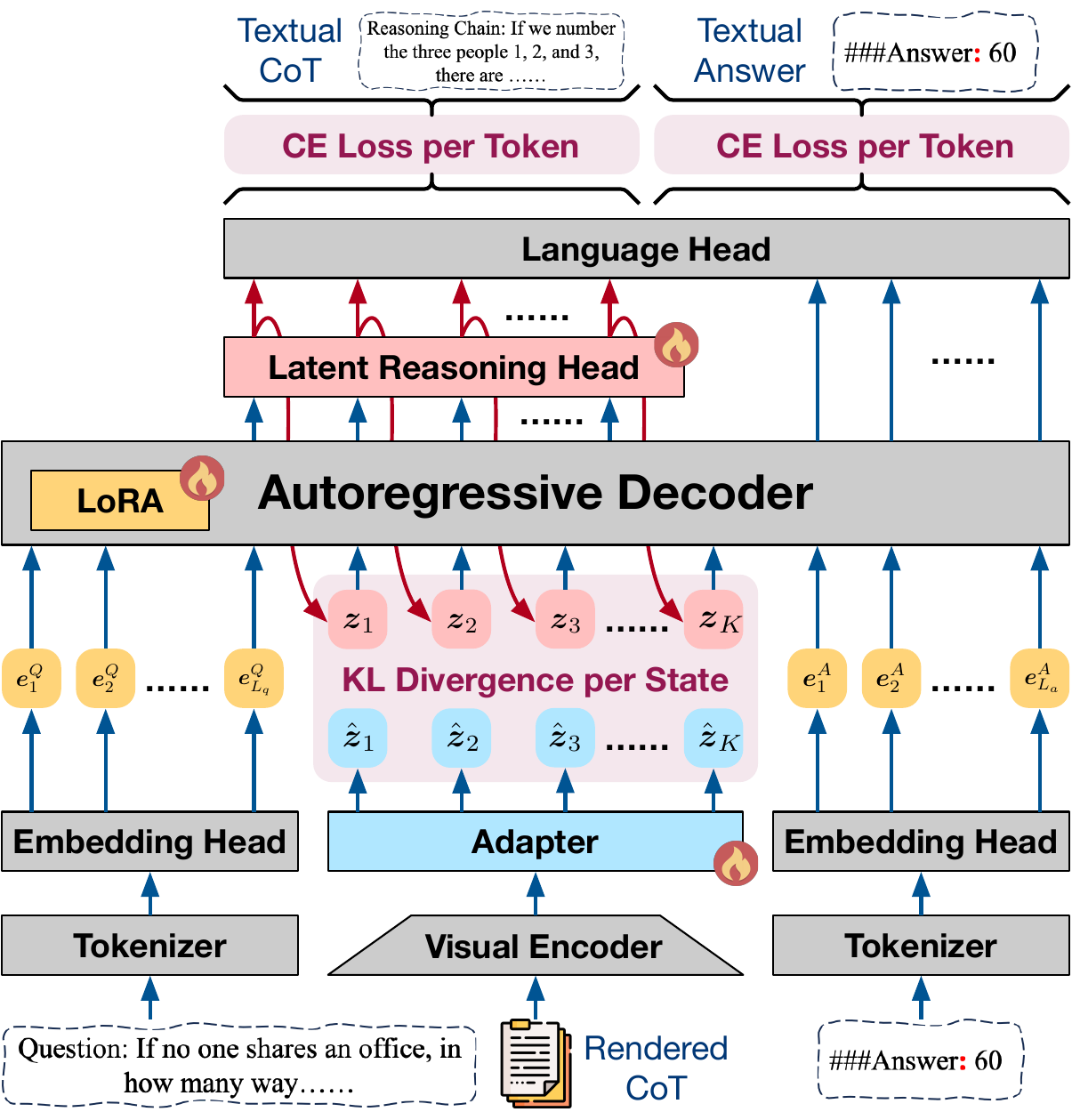} 
    \caption{Illustration of the proposed ReGuLaR. Here, only the latent reasoning head, adapter, and LoRA module are trainable. The blue arrows ``\textcolor{RoyalBlue}{$\rightarrow$}'' indicate deterministic outputs, while the red arrows ``\textcolor{BrickRed}{$\rightarrow$}'' indicate probabilistic outputs achieved by sampling.
    The special token ``\text{\#\#\#}'' triggers the transition from the reasoning process to answer generation during inference.}
    \label{fig: Overview}
    \vspace{-10pt} 
\end{wrapfigure}

Inspired by recent advances~\citep{xing2025vision_vist, wei2025deepseek_OCR} establishing visual representations as compact carriers of textual information, we propose ReGuLaR, a rendered CoT-guided variational latent reasoning method, to implement the VAE framework in~\eqref{eq: learning}. 
As illustrated in Figure~\ref{fig: Overview}, ReGuLaR parametrizes the prior distribution of latent reasoning states based on the visual representation of rendered CoT.
Formally, we can render $K$ segments of the reasoning chain $\mathcal{R}$ into $K$ images and extract their visual representation as follows: For $k=1,...,K$, 
\begin{equation}\label{eq:rendering}
    \begin{aligned}
        & 1)~\text{Rendering:}\quad \mathcal{I}_k = f(\mathcal{R}_k), \\
        & 2)~\text{Embedding:}\quad \bm{v}_k = v(\mathcal{I}_k), \\
        & 3)~\text{Adaptation:}\quad \hat{\bm{z}}_k = g_{\gamma}(\bm{v}_k).
    \end{aligned}
\end{equation}
Here, $f$ is the predefined rendering function, which maps an arbitrary token sequence to an image with a size $H\times W$. 
$v$ is the pretrained visual encoder, which transforms pixel-wise images into visual representations.
We employ the trainable adapter $g_{\gamma}: \mathbb{R}^{d_v} \mapsto \mathbb{R}^{d_h}$ to map visual representations $\bm{V}=[\bm{v}_{1},...,\bm{v}_{K}]$ to the proposed latent reasoning space.

In this study, we directly adopt the optimal rendering configuration identified in Glyph~\citep{cheng2025glyph} for $f$, which maximizes the semantic density.
We implement $v$ as the pretrained visual encoder in DeepSeek-OCR~\citep{wei2025deepseek_OCR} since it has been architecturally optimized for visual-text compression, enabling it to encode high-resolution and text-dense inputs into compact representations with minimal semantic loss.
The adapter $g_{\gamma}$ is instantiated as a multi-layer perception (MLP) with parameters $\gamma$.
Notably, as $f$ and $v$ are frozen in our work, \textbf{we can pre-compute these visual representations offline before training}, significantly reducing computational overhead.

As a result, given the reasoning chain $\mathcal{R}$, we first segment it into $K$ parts\footnote{We regarding each sentence in $\mathcal{R}$ as a segment in Table~\ref{tab:main-results} and analyze the impact of compression rate (i.e., $|\mathcal{R}|/K$) in Figure~\ref{fig: Compression}.} and then model $p_{\gamma}(\cdot|\mathcal{R}_k)$ as a normal distribution $\mathcal{N}(\hat{\bm{z}}_k,\bm{I})$ for $k=1,...,K$, whose mean is determined by~\eqref{eq:rendering} and variance is fixed as an identity matrix.

Accordingly, for $k=1,...,K$, the KL divergence in~\eqref{eq: learning} becomes
\begin{eqnarray}\label{eq:kld}
\begin{aligned}
    \text{KL}[p_{\phi}\|p_{\gamma}]
    =\frac{\|\bm{\mu}_k-\hat{\bm{z}}_k\|_2^2+\|\bm{\sigma}_k\|_2^2}{2}-\log|\text{diag}(\bm{\sigma}_k)|.\!\!
\end{aligned}
\end{eqnarray}

Following the previous work~\citep{colar_tan2025think}, we approximate the KL divergence during training as follows:
\begin{eqnarray}\label{eq:kld2}
\begin{aligned}
    \frac{1}{2}\mathbb{E}_{\bm{\epsilon}\sim \mathcal{N}(\bm{0},\bm{I})}[\|\bm{\mu}_k+\bm{\sigma}_k\odot\bm{\epsilon}-\hat{\bm{z}}_k\|_2^2]-\log|\text{diag}(\bm{\sigma}_k)|.
\end{aligned}
\end{eqnarray}

\textbf{Notably, the LLM trained by ReGuLaR still follows the standard latent reasoning workflow initiated solely by the textual input question:} the model utilizes the trained latent reasoning head to generate latent reasoning states until the special termination token is encountered, which acts as a signal triggering the language head to decode the final answer.
Algorithms~\ref{alg: regular} and~\ref{alg: inference} in Appendix~\ref{app: imple_details} have presented training and inference schemes of ReGuLaR in detail.

\subsection{Advantages over Existing Methods}\label{ssec: advantage}
Unlike the token grouping strategy in CoLaR~\citep{colar_tan2025think}, ReGuLaR renders reasoning chains into images to preserve semantic integrity.
The visual representations of these rendered images provide better regularization than the aggregation of grouped token embeddings, resulting in less information loss. 
As previously noted, the inference phase remains consistent with standard latent reasoning, accepting pure text inputs and imposing no extra computational cost.

Moreover, non-textual elements (e.g., charts, graphs, and diagrams) can also be rendered and encoded alongside text. 
Therefore, in addition to mitigating textual information loss, ReGuLaR natively supports the use of multi-modal information in its latent reasoning processes.
In complex tasks involving multi-modal reasoning information, this advantage enables ReGuLaR not only to outperform existing latent reasoning methods but also to surpass explicit textual CoT.

\section{Experiments}
\subsection{Experimental Setup}
\textbf{Datasets.} 
Following previous work~\citep{colar_tan2025think}, we primarily train and evaluate models on GSM8K-Aug~\citep{deng2023implicit}, and additionally evaluate trained models using three out-of-domain math reasoning datasets: GSM-Hard~\citep{gao2023pal}, SVAMP~\citep{patel2021nlp}, and MultiArith~\citep{roy2015solving}. 
Meanwhile, we also train and evaluate on GSM8K-Aug-NL, the variant of GSM8K-Aug that preserves natural language explanations within reasoning processes, to explore their extreme compression capabilities. 
Additionally, we conduct experiments on AQUA-RAT~\citep{ling2017program} and MATH~\citep{hendrycks2021measuring} to verify their performance in more challenging problems.

\textbf{Baselines.} 
We employ various latent reasoning methods as baselines, including iCoT~\citep{deng2024explicit}, CODI~\citep{shen2025codi}, Coconut~\citep{coconut_hao2024training}, and CoLaR~\citep{colar_tan2025think}.
Specifically, all these baselines are implemented following their default configurations within the unified framework provided by the state-of-the-art method CoLaR, thereby ensuring fairness and consistency.

\textbf{Evaluation.} 
We adopt standard evaluation metrics widely used in this domain~\citep{coconut_hao2024training} to assess performance, including $i)$ Accuracy (Acc.), which evaluates reasoning effectiveness by calculating the percentage of correctly predicted answers; and $ii)$ Reasoning Length (\#~L), which evaluates reasoning efficiency by calculating the number of reasoning steps per question. 
Specifically, all these evaluations are repeated over five independent runs with different random seeds, thereby ensuring statistical reliability.

\textbf{Implementation.} 
Unless otherwise specified, we employ LLaMA-3.2-1B-Instruct (\textbf{LLaMA-1B}~\citep{grattafiori2024llama}) as the LLM backbone.
Specifically, following established baselines, we keep the LLM backbone frozen and exclusively optimize LoRA~\citep{hulora} modules configured with $r=128$ and $\alpha=32$.

More details about datasets, baselines, and implementation have been provided in Appendix~\ref{app: details}.

\begin{table*}[t]
\centering
\caption{Performance comparison on four math reasoning datasets using LLaMA-3.2-1B-Instruct as the LLM backbone. We report the averaged number and 95\% confidence interval ($\pm$) on Accuracy (Acc. \%) and Reasoning Length ($\#$ L).}
\label{tab:main-results}
\setlength{\tabcolsep}{3.85pt}
\small{
\begin{tabular}{lcccccccccc}
\toprule
\multirow{2}{*}[-0.55ex]{Method} & \multicolumn{2}{c}{GSM8K-Aug} & \multicolumn{2}{c}{GSM-Hard} & \multicolumn{2}{c}{SVAMP} & \multicolumn{2}{c}{MultiArith} & \multicolumn{2}{c}{Average}\\
\cmidrule(lr){2-3} \cmidrule(lr){4-5} \cmidrule(lr){6-7} \cmidrule(lr){8-9} \cmidrule(lr){10-11}
& \multicolumn{1}{c}{Acc.} & \multicolumn{1}{c}{\#~L} & \multicolumn{1}{c}{Acc.} & \multicolumn{1}{c}{\#~L} & \multicolumn{1}{c}{Acc.} & \multicolumn{1}{c}{\#~L} & \multicolumn{1}{c}{Acc.} & \multicolumn{1}{c}{\#~L} & \multicolumn{1}{c}{Acc.} & \multicolumn{1}{c}{\#~L} \\
\midrule
iCoT & $19.8_{\pm0.23}$ & $0.00_{\pm0.00}$ & $3.87_{\pm0.16}$ & $0.00_{\pm0.00}$ & $36.4_{\pm0.51}$ & $0.00_{\pm0.00}$ & $38.2_{\pm0.66}$ & $0.00_{\pm0.00}$ & 24.6 & 0.00 \\
CODI & $13.3_{\pm0.62}$ & $6.00_{\pm0.00}$ & $2.97_{\pm0.24}$ & $6.00_{\pm0.00}$ & $21.7_{\pm0.73}$ & $6.00_{\pm0.00}$ & $19.2_{\pm0.83}$ & $6.00_{\pm0.00}$ & 14.3 & 6.00 \\
Coconut & $20.5_{\pm0.68}$ & $6.00_{\pm0.00}$ & $4.86_{\pm0.30}$ & $6.00_{\pm0.00}$ & $39.8_{\pm0.71}$ & $6.00_{\pm0.00}$ & $41.4_{\pm0.69}$ & $6.00_{\pm0.00}$ & 26.6 & 6.00 \\
CoLaR* & $26.6_{\pm0.18}$ & $5.63_{\pm0.01}$ & $6.23_{\pm0.14}$ & $7.01_{\pm0.05}$ & $47.1_{\pm0.30}$ & $2.96_{\pm0.02}$ & $87.0_{\pm0.21}$ & $3.23_{\pm0.01}$ & 41.7 & 4.70 \\
\rowcolor{highlightred}
ReGuLaR & $\textbf{34.9}_{\pm0.26}$ & $\textbf{3.69}_{\pm0.21}$ & $\textbf{8.27}_{\pm0.14}$ & $\textbf{4.12}_{\pm0.48}$ & $\textbf{50.1}_{\pm0.39}$ & $\textbf{2.02}_{\pm0.18}$ & $\textbf{89.2}_{\pm0.27}$ & $\textbf{2.28}_{\pm0.27}$ & \textbf{45.6} & \textbf{3.03} \\ 
\bottomrule
\end{tabular}
}
\par \vspace{-2.5pt}
\footnotesize{*Results here utilize the default maximum compression rate of 5, while comparison across varying rates is provided in Figure~\ref{fig: Compression}}
\end{table*}

\begin{figure*}[t]
    \centering 
    \begin{subfigure}{0.495\textwidth}
        \centering
        \includegraphics[width=\linewidth]{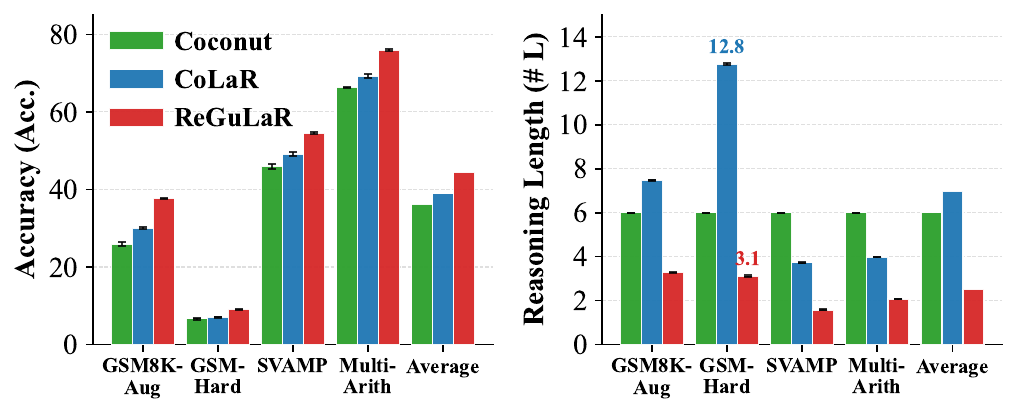} 
        \caption{Generalizability Analysis}
        \label{fig: Generalization}
    \end{subfigure}
    \hfill
    \begin{subfigure}{0.495\textwidth}
        \centering
        \includegraphics[width=\linewidth]{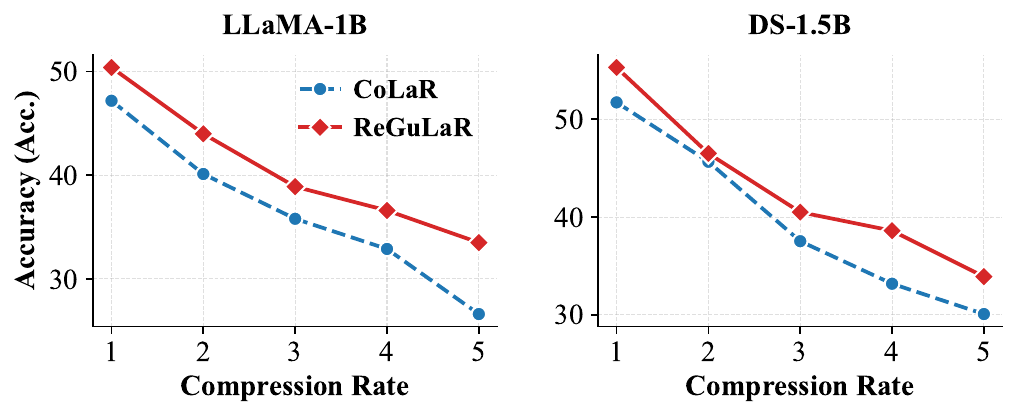} 
        \caption{Compression Analysis}
        \label{fig: Compression}
    \end{subfigure}
    \caption{(a) Generalizability analysis using DeepSeek-R1-Distill-Qwen-1.5B as the LLM backbone, where the left panel reports Accuracy and the right panel reports Reasoning Length. (b) Compression Analysis on the GSM8K-Aug dataset using LLaMA-3.2-1B-Instruct (left) and DeepSeek-R1-Distill-Qwen-1.5B (right) as the LLM backbone, where the compression rate represents the number of explicit reasoning tokens corresponding to a single latent reasoning state. }
\end{figure*}

\subsection{Main Results}
\textbf{Performance Comparison.}
Table~\ref{tab:main-results} presents the results of various methods on four math reasoning datasets, where ReGuLaR achieves state-of-the-art performance.
Specifically, compared with the strongest baseline CoLaR, ReGuLaR consistently delivers substantial accuracy gains across all datasets while simultaneously reducing the average reasoning length by approximately 35\% (from 4.70 to 3.03).
The results suggest that ReGuLaR successfully compresses the reasoning chain into a more compact and informative latent reasoning state sequence, underscoring both its computational efficiency and reasoning effectiveness.

\textbf{Generalizability Analysis.}
To demonstrate the generalizability of ReGuLaR to different LLM backbones, we replace the underlying model (i.e., LLaMA-3.2-1B-Instruct) with DeepSeek-R1-Distill-Qwen-1.5B.
As shown in Figure~\ref{fig: Generalization}, ReGuLaR consistently maintains its superiority across all datasets, achieving the highest accuracy with the shortest reasoning length.
Especially on the GSM-Hard dataset, while the strongest baseline CoLaR requires an average of 12.8 reasoning steps, ReGuLaR achieves higher accuracy with only 3.1 steps.

\textbf{Compression Analysis.} 
While the strongest baseline CoLaR employs token embedding compression and ReGuLaR leverages visual-text compression, both map a sequence of explicit reasoning tokens to a single latent reasoning state.
Given this commonality, we assess their performance under identical compression rates (i.e., the number of tokens condensed into one latent reasoning state).
As shown in Figure~\ref{fig: Compression}, although accuracy naturally decreases with higher compression rates, ReGuLaR consistently outperforms CoLaR across all settings on both LLM backbones, verifying its advantage in preserving semantic information.

\begin{figure*}[t]
    \centering 
    \includegraphics[width=\textwidth]{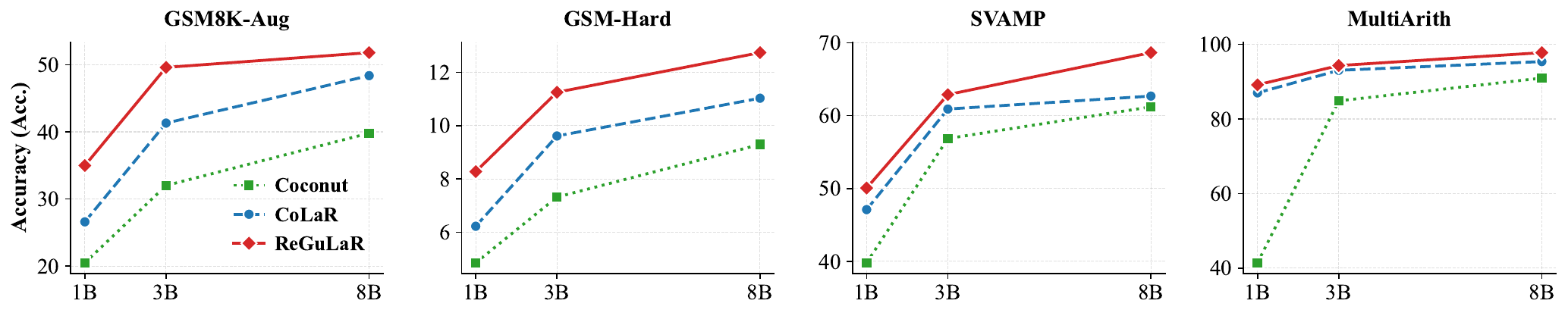} 
    \caption{Scalability analysis across varying model sizes, where we employ LLaMA-3.2 (1B, 3B) and LLaMA-3.1 (8B) Instruct variants as the LLM backbones. Comprehensive results, including reasoning length, are provided in Figure~\ref{fig: Full_Scalability}.}
    \label{fig: Scalability}
\end{figure*}

\textbf{Scalability Analysis.}
To evaluate the scaling potential of ReGuLaR, we conduct experiments across varying model sizes within the LLaMA-3 family, ranging from 1B to 8B (i.e., instruct variants of LLaMA-3.2 1B/3B and LLaMA-3.1 8B).
As shown in Figure~\ref{fig: Scalability}, ReGuLaR demonstrates strong positive scaling behavior, consistently maintaining a significant performance margin over the top-performing baselines (CoLaR and Coconut) across all model scales and datasets, demonstrating its seamless scalability and potential for broader application in large-scale foundation models.

More experimental results, including ablation studies, have been provided in Appendix~\ref{app: results}.

\begin{table*}[t]
\centering
\caption{Extreme compression performance of ReGuLaR, where CoLaR follows its default configuration as a reference.}
\label{tab:extreme-compression}
\setlength{\tabcolsep}{1.85pt}
\small{
\begin{tabular}{clcccccccccc}
\toprule
\multirow{2}{*}[-0.55ex]{Dataset} &\multirow{2}{*}[-0.55ex]{Method} & \multicolumn{2}{c}{LLaMA-1B} & \multicolumn{2}{c}{LLaMA-3B} & \multicolumn{2}{c}{LLaMA-8B} & \multicolumn{2}{c}{DS-1.5B} & \multicolumn{2}{c}{Average} \\
\cmidrule(lr){3-4} \cmidrule(lr){5-6} \cmidrule(lr){7-8} \cmidrule(lr){9-10} \cmidrule(lr){11-12} 
& & \multicolumn{1}{c}{Acc.} & \multicolumn{1}{c}{\#~L} & \multicolumn{1}{c}{Acc.} & \multicolumn{1}{c}{\#~L} & \multicolumn{1}{c}{Acc.} & \multicolumn{1}{c}{\#~L} & \multicolumn{1}{c}{Acc.} & \multicolumn{1}{c}{\#~L}  & \multicolumn{1}{c}{Acc.} & \multicolumn{1}{c}{\#~L} \\
\midrule
\cellcolor{white} GSM8K- & CoLaR & $18.7_{\pm0.34}$ & $13.1_{\pm0.04}$  & $31.0_{\pm0.26}$ & $14.8_{\pm0.11}$ & $31.7_{\pm0.21}$ & $13.1_{\pm0.07}$  & $16.4_{\pm0.23}$ & $15.4_{\pm0.03}$ & 24.4 & 14.1 \\
\rowcolor{highlightred}
\cellcolor{white} Aug-NL
& ReGuLaR &  $\textbf{20.2}_{\pm0.71}$ & $\textbf{1.00}_{\pm0.00}$ &  $\textbf{32.6}_{\pm0.43}$ & $\textbf{1.00}_{\pm0.00}$ &  $\textbf{38.6}_{\pm0.33}$ & $\textbf{1.00}_{\pm0.00}$ & 
$\textbf{31.3}_{\pm0.09}$ & $\textbf{1.00}_{\pm0.00}$ & \textbf{30.7} & \textbf{1.00} \\ 
\midrule
AQUA-& CoLaR & $24.2_{\pm0.37}$ & $19.4_{\pm0.31}$  & $31.7_{\pm1.27}$ & $28.5_{\pm0.66}$ & $35.8_{\pm0.90}$ & $20.5_{\pm1.66}$  & $33.2_{\pm1.04}$ & $26.7_{\pm0.54}$ & 31.2 & 23.8 \\
\rowcolor{highlightred}
\cellcolor{white}{RAT}
& ReGuLaR &  $\textbf{37.1}_{\pm0.61}$ & $\textbf{1.00}_{\pm0.00}$ &  $\textbf{39.7}_{\pm0.50}$ & $\textbf{1.00}_{\pm0.00}$ &  $\textbf{41.2}_{\pm0.34}$ & $\textbf{1.00}_{\pm0.00}$ & 
$\textbf{41.0}_{\pm0.48}$ & $\textbf{1.00}_{\pm0.00}$ & \textbf{39.8} & \textbf{1.00} \\ 
\midrule
& CoLaR & $3.65_{\pm0.13}$ & $58.1_{\pm0.29}$  & $8.03_{\pm0.25}$ & $60.4_{\pm0.19}$ & $9.06_{\pm0.20}$ & $67.7_{\pm0.59}$  & $10.3_{\pm0.22}$ & $62.6_{\pm0.21}$ & 7.76 & 62.2 \\
\rowcolor{highlightred}
\cellcolor{white}\multirow{-2}{*}{MATH}
& ReGuLaR &  $\textbf{6.62}_{\pm0.18}$ & $\textbf{1.00}_{\pm0.00}$ &  $\textbf{11.8}_{\pm0.03}$ & $\textbf{1.00}_{\pm0.00}$ &  $\textbf{13.9}_{\pm0.10}$ & $\textbf{1.00}_{\pm0.00}$ & 
$\textbf{15.6}_{\pm0.14}$ & $\textbf{1.00}_{\pm0.00}$ & \textbf{11.9} & \textbf{1.00} \\ 
\bottomrule
\end{tabular}
}
\end{table*}

\begin{table*}[t]
\centering
\caption{Performance comparison on the molecular captioning task. 
We report the averaged number and 95\% confidence interval ($\pm$) on BLEU, METEOR, ROUGE, and Reasoning Length (\# L) via five independent evaluations. Note that "w/o 2D" denotes the variant of ReGuLaR trained with the original textual reasoning chains for ablation.}
\setlength{\tabcolsep}{2pt}
\label{tab: molecular captioning}
\resizebox{\textwidth}{!}{
\begin{tabular}{l|l|ccccccc}
\toprule
\multicolumn{2}{l|}{Method} & BLEU-2$\uparrow$ & BLEU-4$\uparrow$ & METEOR$\uparrow$ & ROUGE-1$\uparrow$ & ROUGE-2$\uparrow$ & ROUGE-L$\uparrow$ & \# L \\
\midrule
\multicolumn{2}{l|}{GPT-4o}      & 0.1093 & 0.0313 & 0.1682 & 0.2491 & 0.0778 & 0.1875 & / \\
\multicolumn{2}{l|}{DeepSeek-R1} & 0.1011 & 0.0297 & 0.2178 & 0.2506 & 0.0709 & 0.1722 & / \\
\midrule
&CoT     & $0.2828_{\pm 0.002}$ & $0.1804_{\pm 0.002}$ & $0.3778_{\pm 0.001}$ & $0.4632_{\pm 0.002}$ & $0.2675_{\pm 0.003}$ & $0.3914_{\pm 0.001}$ & $314.825_{\pm 4.143}$ \\
&CoLaR   & $0.0995_{\pm 0.002}$ & $0.0400_{\pm 0.001}$ & $0.1774_{\pm 0.002}$ & $0.2306_{\pm 0.004}$ & $0.0911_{\pm 0.002}$ & $0.1900_{\pm 0.003}$ & $212.347_{\pm 0.660}$\\
\rowcolor{highlightred}
\cellcolor{white}\multirow{-4}{*}{}
&ReGuLaR   & $\textbf{0.3673}_{\pm 0.002}$ & $\textbf{0.2692}_{\pm 0.002}$ & $\textbf{0.4593}_{\pm 0.002}$ & $\textbf{0.5319}_{\pm 0.002}$ & $\textbf{0.3502}_{\pm 0.002}$ & $\textbf{0.4635}_{\pm 0.002}$ & $\phantom{00}\textbf{1.000}_{\pm 0.000}$ \\
\rowcolor{graybg}
\cellcolor{white}\multirow{-4}{*}{LLaMA-1B}
& \quad w/o 2D & $0.2872_{\pm 0.002}$ &  $0.1845_{\pm 0.002}$ & $0.3777_{\pm 0.001}$ & $0.4678_{\pm 0.001}$ & $0.2716_{\pm 0.001}$ & $0.3962_{\pm 0.000}$ & $\phantom{00}{1.000}_{\pm 0.000}$ \\
\midrule
&CoT     & $0.3167_{\pm 0.003}$ & $0.2146_{\pm 0.003}$ & $0.4117_{\pm 0.003}$ & $0.4948_{\pm 0.001}$ & $0.3011_{\pm 0.003}$ & $0.4230_{\pm 0.002}$ & $316.829_{\pm 3.502}$ \\
&CoLaR   & $0.1734_{\pm 0.002}$ & $0.0861_{\pm 0.002}$ & $0.2487_{\pm 0.003}$ & $0.3381_{\pm 0.002}$ & $0.1576_{\pm 0.001}$ & $0.2840_{\pm 0.002}$ & $200.267_{\pm 5.437}$ \\
\rowcolor{highlightred}
\cellcolor{white}\multirow{-4}{*}{}
&ReGuLaR   & $\textbf{0.4329}_{\pm 0.002}$ & $\textbf{0.3394}_{\pm 0.002}$ & $\textbf{0.5158}_{\pm 0.002}$ & $\textbf{0.5860}_{\pm 0.002}$ & $\textbf{0.4164}_{\pm 0.003}$ & $\textbf{0.5208}_{\pm 0.002}$ & $\phantom{00}\textbf{1.000}_{\pm 0.000}$  \\
\rowcolor{graybg}
\cellcolor{white}\multirow{-4}{*}{LLaMA-3B}
& \quad w/o 2D & $0.3182_{\pm 0.003}$ &  $0.2119_{\pm 0.003}$ & $0.4172_{\pm 0.002}$ & $0.4940_{\pm 0.002}$ & $0.2961_{\pm 0.003}$ & $0.4184_{\pm 0.002}$ & $\phantom{00}{1.000}_{\pm 0.000}$ \\
\midrule
&CoT & $0.3410_{\pm 0.001}$ & $0.2378_{\pm 0.001}$ & $0.4377_{\pm 0.001}$ & $0.5138_{\pm 0.001}$ & $0.3212_{\pm 0.001}$ & $0.4403_{\pm 0.001}$ & $295.736_{\pm 4.105}$ \\
&CoLaR & $0.2031_{\pm 0.003}$ & $0.1385_{\pm 0.002}$ & $0.2803_{\pm 0.003}$ & $0.3883_{\pm 0.001}$ & $0.2340_{\pm 0.002}$ & $0.3449_{\pm 0.002}$ & $163.655_{\pm 3.981}$ \\
\rowcolor{highlightred}
\cellcolor{white}\multirow{-4}{*}{}
&ReGuLaR   & $\textbf{0.4610}_{\pm 0.001}$ & $\textbf{0.3691}_{\pm 0.001}$ & $\textbf{0.5479}_{\pm 0.002}$ & $\textbf{0.6094}_{\pm 0.001}$ & $\textbf{0.4428}_{\pm 0.001}$ & $\textbf{0.5439}_{\pm 0.001}$ & $\phantom{00}\textbf{1.000}_{\pm 0.000}$  \\
\rowcolor{graybg}
\cellcolor{white}\multirow{-4}{*}{LLaMA-8B}
& \quad w/o 2D & $0.3403_{\pm 0.002}$ &  $0.2364_{\pm 0.001}$ & $0.4394_{\pm 0.002}$ & $0.5132_{\pm 0.002}$ & $0.3187_{\pm 0.002}$ & $0.4367_{\pm 0.003}$ & $\phantom{00}{1.000}_{\pm 0.000}$ \\
\midrule
&CoT     & $0.2682_{\pm 0.002}$ & $0.1659_{\pm 0.002}$ & $0.3637_{\pm 0.002}$ & $0.4469_{\pm 0.001}$ & $0.2523_{\pm 0.001}$ & $0.3774_{\pm 0.001}$ & $344.067_{\pm 6.366}$ \\
&CoLaR   & $0.0968_{\pm 0.002}$ & $0.0516_{\pm 0.001}$ & $0.1644_{\pm 0.002}$ & $0.2149_{\pm 0.004}$ & $0.0916_{\pm 0.001}$ & $0.1801_{\pm 0.003}$ & $580.558_{\pm 9.779}$ \\
\rowcolor{highlightred}
\cellcolor{white}\multirow{-4}{*}{}
&ReGuLaR   & $\textbf{0.3536}_{\pm 0.002}$ & $\textbf{0.2545}_{\pm 0.001}$ & $\textbf{0.4397}_{\pm 0.001}$ & $\textbf{0.5187}_{\pm 0.002}$ & $\textbf{0.3347}_{\pm 0.001}$ & $\textbf{0.4514}_{\pm 0.001}$ & $\phantom{00}\textbf{1.000}_{\pm 0.000}$ \\
\rowcolor{graybg}
\cellcolor{white}\multirow{-4}{*}{DS-1.5B}
& \quad w/o 2D & $0.2672_{\pm 0.002}$ &  $0.1640_{\pm 0.002}$ & $0.3764_{\pm 0.002}$ & $0.4645_{\pm 0.002}$ & $0.2499_{\pm 0.002}$ & $0.3476_{\pm 0.003}$ & $\phantom{00}{1.000}_{\pm 0.000}$ \\
\bottomrule
\end{tabular}}
\end{table*}

\subsection{Extreme Compression with ReGuLaR}\label{ssec: extreme}
As expressed in~\eqref{eq:rendering}, the reasoning chain $\mathcal{R}$ is decomposed into $K$ segments, which are subsequently rendered into images to yield $K$ corresponding latent reasoning states during training, thereby impacting information preservation and compression efficiency.
In Table~\ref{tab:main-results}, we follow the natural linguistic partition, treating each sentence as a segment.
In this section, we further investigate the limit of model performance by introducing an extreme compression setting.
Specifically, we directly render the entire reasoning chain into a single image, compressing all reasoning information into one latent reasoning state (i.e., $K=1$).
We conduct this experiment on the GSM8K-Aug-NL dataset, which preserves natural language explanations within the reasoning process, inherently yielding relatively long reasoning chains.
The AQUA-RAT and MATH datasets are also incorporated to verify the performance on more challenging problems.

Table~\ref{tab:extreme-compression} summarizes the performance of ReGuLaR under the extreme compression setting.
Specifically, despite being constrained to a single latent reasoning step, ReGuLaR still outperforms the strongest baseline CoLaR across all model scales and datasets.
This advantage is particularly evident on the MATH dataset, where ReGuLaR improves average accuracy from $7.76\%$ to $11.9\%$ while reducing reasoning length from $62.2$ to $1.00$, thereby underscoring its superior compression capability in complex reasoning scenarios. 

\section{Latent Reasoning Beyond Textual Domain}\label{sec:multimodal}
As discussed in Section~\ref{ssec: advantage}, non-textual elements can also be rendered alongside text, making ReGuLaR support multi-modality within latent reasoning while maintaining the standard textual I/O interface.
In this section, we conduct experiments to investigate the efficacy of this extended capability.

\textbf{Dataset.} 
Unlike existing multi-modal datasets that rely on image-based inputs and textual reasoning chains, we require datasets that maintain purely textual I/O while integrating multi-modal reasoning chains as intermediate bridges.
To this end, we adopt the \textbf{molecule captioning benchmark} from MolReasoner~\citep{zhao2025molreasoner}, which requires the LLM to generate natural language descriptions of the given molecules (represented by SELFIES strings).
In particular, while the original dataset only provides textual reasoning chains, we utilize RDKit~\citep{landrum2013rdkit} to generate the corresponding 2D molecular graphs and combine them with the original textual reasoning chains, thereby constructing multi-modal reasoning chains.

\textbf{Baselines and Evaluation.} 
Consistent with Section~\ref{ssec: extreme}, we also render the entire multi-modal reasoning chain into a single image to train ReGuLaR.
The baselines include CoT and CoLaR, both of which apply the original textual reasoning chains for training due to their inherent lack of multi-modal support. 
The model performance is evaluated by the widely used \textbf{BLEU}, \textbf{METEOR}, and \textbf{ROUGE} metrics, which quantify the $n$-gram overlap and semantic similarity between the generated and reference captions. 

\textbf{Results.}
As presented in Table~\ref{tab: molecular captioning}, ReGuLaR achieves state-of-the-art performance across all metrics and backbones.
Specifically, despite being constrained to a single latent reasoning step, ReGuLaR significantly outperforms not only the strongest latent reasoning baseline CoLaR, but also the explicit CoT method, both of which apply hundreds of reasoning steps. 
Notably, although CoT and CoLaR are trained using the original textual reasoning chains due to their inherent lack of multi-modal support, the comparison between our method and these two methods is fair to some extent because the 2D graphs in the multi-modal reasoning chains only convert the data format, which do not provide any additional information.
To ensure a strictly fair comparison, we construct an ablation setting (denoted as ``w/o 2D'') where only textual reasoning chains are rendered.
Even in this setting, ReGuLaR maintains performance comparable to CoT while drastically reducing the reasoning steps. 
These results further validate the extreme compression capability of ReGuLaR and highlight its unique advantage of unifying textual and non-textual elements for comprehensive reasoning.

\section{Conclusion and Future Work}
In this paper, we propose a new and insightful latent reasoning paradigm that models latent reasoning within the VAE framework and learns it guided by rendered CoT.
Our method significantly outperforms existing latent reasoning methods across both computational efficiency and reasoning ability, and even surpasses explicit CoT through supporting multi-modal latent reasoning.

\textbf{Future work.}
Currently, standard benchmarks like GSM8K and GSM8K-Aug may limit the assessment of advanced reasoning capabilities because they have limited data sizes and overly simple reasoning chains. 
We plan to address this gap by developing a large-scale and high-quality reasoning dataset to evaluate latent reasoning methods in more demanding settings.
In addition, we will further explore latent reasoning and study whether and how it can outperform explicit CoT in theory.

\bibliography{refs}

\newpage
\appendix
\section*{Appendix}
\section{More Experimental Details}\label{app: details}
\subsection{Datasets}
Following previous work~\citep{colar_tan2025think}, we primarily train and evaluate our method on the GSM8K-Aug dataset~\citep{deng2023implicit}, and additionally evaluate trained models using three out-of-domain math reasoning datasets: GSM-Hard~\citep{gao2023pal}, SVAMP~\citep{patel2021nlp}, and MultiArith~\citep{roy2015solving}. 
Meanwhile, we also train and evaluate on the GSM8K-Aug-NL dataset, a variant of the GSM8K-Aug dataset that preserves natural language explanations, to demonstrate the extreme compression capability. 
In addition, we extend our experiments to the AQUA-RAT dataset~\citep{ling2017program} and MATH dataset~\citep{hendrycks2021measuring} to verify the performance on more challenging problems.

Here, the detailed description of the above datasets is provided below:
\begin{itemize}
    \item \textbf{GSM8K-Aug}~\citep{deng2023implicit}: 
    This dataset is an augmented version of the GSM8K dataset~\citep{cobbe2021gsm8k}, constructed by prompting GPT-4~\citep{achiam2023gpt} to extend the original training set to 385k samples.
    In particular, it eliminates natural language descriptions within the reasoning process, formalizing reasoning steps as mathematical expressions only.
    \item \textbf{GSM8K-Aug-NL}~\citep{deng2023implicit}: 
    Similar to the GSM8K-Aug dataset, this dataset is also constructed by prompting GPT-4~\citep{achiam2023gpt} to extend the original training set of the GSM8K dataset to 385k samples.
    The distinguishing feature is that GSM8K-Aug-NL preserves natural language explanations within the reasoning process, formalizing reasoning steps as natural language sentences.
    Consequently, compared to GSM8K-Aug, this dataset exhibits longer reasoning chains, with a reasoning style that more closely resembles verbose CoTs.
    \item \textbf{GSM8K-Hard}~\citep{gao2023pal}:
    This dataset is the harder version of the GSM8K dataset, constructed by replacing the numbers in the original test set with larger numbers that are less common.
    In particular, this dataset serves as the out-of-domain dataset in our experiments.
    \item \textbf{SVAMP}~\citep{patel2021nlp}: 
    This dataset comprises 1,000 elementary-level math word problems, designed to evaluate the robustness of models in solving fundamental mathematical problems.
    In particular, this dataset serves as the out-of-domain dataset in our experiments.
    \item \textbf{MultiArith}~\citep{roy2015solving}:
    This dataset comprises 600 multi-step arithmetic problems, designed to evaluate the capability of models in handling tasks that require multi-step reasoning.
    In particular, this dataset serves as the out-of-domain dataset in our experiments.
    \item \textbf{AQUA-RAT}~\citep{ling2017program}: This dataset comprises about 100,000 algebraic word problems, where each problem is equipped with a step-by-step natural language explanation that details the logical derivation leading to the answer.  
    \item \textbf{MATH}~\citep{hendrycks2021measuring}: This dataset comprises 12,500 highly challenging mathematics competition problems, where each problem is equipped with a comprehensive step-by-step solution. 
\end{itemize}

\subsection{Baselines}
We employ various respective latent-based methods as baselines, including iCoT~\citep{deng2024explicit}, CODI~\citep{shen2025codi}, Coconut~\citep{coconut_hao2024training}, and CoLaR~\citep{colar_tan2025think}.

Here, the detailed description of the above baselines is provided below:
\begin{itemize}
    \item \textbf{iCoT}~\citep{deng2024explicit}: This baseline gradually removes the intermediate reasoning steps during finetuning, thereby internalizing the reasoning process while maintaining high performance.
    \item \textbf{CODI}~\citep{shen2025codi}: This baseline directly employs self-distillation to align the hidden activations of latent thoughts with CoT trajectories, thereby transferring reasoning capabilities into latent space.
    \item \textbf{Coconut}~\citep{coconut_hao2024training}: This baseline recursively utilizes the last hidden state of LLMs as latent thought, thereby functioning as the next input embedding to drive subsequent reasoning.
    \item \textbf{CoLaR}~\citep{colar_tan2025think}: This baseline dynamically compresses embeddings of reasoning tokens and autoregressively predicts compressed embeddings, thereby supporting flexible reasoning lengths. 
\end{itemize}

\begin{table}[t]
    \centering
    \caption{Detailed rendering configuration used in our experiments.}
    \label{tab:rendering_config}{
    \begin{tabular}{l|c|c}
        \toprule
        \textbf{Parameter} & \textbf{Value} & \textbf{Description} \\
        \midrule
        \multicolumn{3}{l}{\textit{Canvas \& Layout}} \\
        \midrule
        Page Size & $595 \times 842$ & Standard A4 dimension (points) \\
        DPI & 72 & Screen resolution density \\
        Margins (X, Y) & 10, 10 & Minimal padding to maximize content area \\
        Background Color & \#FFFFFF & White background \\
        Auto Crop & True & Crop to content bounding box \\
        \midrule
        \multicolumn{3}{l}{\textit{Typography}} \\
        \midrule
        Font Family & Verdana & Sans-serif font for optical clarity \\
        Font Size & 9 & Compact size for high density \\
        Line Height & 10 & Tight vertical spacing \\
        Font Color & \#000000 & Black text for high contrast \\
        Alignment & LEFT & Standard left-aligned text \\
        \midrule
        \multicolumn{3}{l}{\textit{Spacing \& Indentation}} \\
        \midrule
        Indent (First/Left/Right) & 0 & No indentation \\
        Spacing (Before/After) & 0 & No paragraph spacing \\
        Border Width & 0 & Borderless rendering \\
        \bottomrule
    \end{tabular}
    }
\end{table}

\subsection{Implementation}\label{app: imple_details}
\textbf{Rendering Configuration.}
As expressed in~\eqref{eq:rendering}, the rendering function $\Phi$ is parameterized by a specific configuration vector $\theta^*$ to map an arbitrary token sequence to an image.
To ensure consistent visual encoding and maximize the semantic density, we directly adopt the optimal rendering configuration identified in Glyph~\citep{cheng2025glyph}.
Specifically, we utilize the \texttt{Verdana} font family and set a tight layout to compact the logical topology.
The detailed rendering specifications are summarized in Table~\ref{tab:rendering_config}.

\textbf{Visual Encoder.}
As expressed in~\eqref{eq:rendering}, rendered images obtained from the preceding stage are mapped into continuous space, transforming pixel-based inputs into visual-semantic vectors.
To encode high-resolution and text-dense inputs into compact representations with minimal semantic loss, we adopt the trained visual encoder from DeepSeek-OCR~\citep{wei2025deepseek_OCR}.
Specifically, this visual encoder integrates a SAM-Base backbone (80M) for fine-grained local perception (via window attention) and a CLIP-Large backbone (300M) for high-level semantic extraction (via global attention), bridged by a $16\times$ convolutional compressor.
Besides, it has been trained via the generative objective on the massive corpus of diverse optical data (including multilingual documents and synthesized charts/formulas), making it align perfectly with our requirements.
In our experiments, we keep this visual encoder frozen and utilize the standard Tiny mode (resolution $512 \times 512$).
This configuration initially maps each rendered image into a sequence of 64 visual tokens, which are subsequently aggregated via mean pooling to derive a single and compact visual-semantic representation.

\begin{algorithm}[H]
\caption{Training Scheme of ReGuLaR}
\label{alg: regular}
\begin{algorithmic}[1]
\REQUIRE Training dataset $\mathcal{D} = \{(\mathcal{Q}, \mathcal{R}, \mathcal{A})\}$, Rendering function $f$, Visual Encoder $v$, Adapter $g_{\gamma}$, LLM with parameters $\theta'$=\{$\phi$, $\psi$\},
\STATE \textbf{Stage 1: Offline Pre-computation}
\FOR{each sample $(\mathcal{Q}, \mathcal{R}, \mathcal{A}) \in \mathcal{D}$}
    \STATE Divide reasoning chain $\mathcal{R}$ into $K$ segments $\{\mathcal{R}_1, ..., \mathcal{R}_K\}$.
    \FOR{$k=1$ to $K$}
        \STATE Render segment to image: $\mathcal{I}_k \leftarrow f(\mathcal{R}_k)$ 
        \STATE Extract visual representation: $\bm{v}_k \leftarrow v(\mathcal{I}_k)$
    \ENDFOR
    \STATE Store pre-computed representations $\bm{V} = [\bm{v}_{1},...,\bm{v}_{K}]$.
\ENDFOR

\STATE \textbf{Stage 2: Training}
\WHILE{not converged}
    \STATE Sample batch of $(\mathcal{Q}, \mathcal{R}, \mathcal{A}, \bm{V})$ from $\mathcal{D}$.
    \STATE Initialize total loss $\mathcal{L}_{total} \leftarrow 0$.
    \STATE Initialize latent reasoning state history $\bm{Z}_{<1} \leftarrow \emptyset$.
    
    \FOR{$k=1$ to $K$}
        \STATE \textcolor{blue}{\textit{// 1. Construct Prior }}
        \STATE Compute prior mean via adapter: $\hat{\bm{z}}_k \leftarrow g_{\gamma}(\bm{v}_k)$
        
        \STATE \textcolor{blue}{\textit{// 2. Sample Posterior }} \textcolor{red}{\textit{(corresponds to~\eqref{eq:posterior})}}
        \STATE Predict posterior parameters: $\bm{\mu}_k, \log\bm{\sigma}_k \leftarrow p_{\phi}(\mathcal{Q}, \bm{Z}_{<k})$
        \STATE Sample latent reasoning state: $\bm{\epsilon} \sim \mathcal{N}(\bm{0}, \bm{I})$, $\bm{z}_k \leftarrow \bm{\mu}_k + \bm{\sigma}_k \odot \bm{\epsilon}$
        \STATE Update history: $\bm{Z}_{<k+1} \leftarrow \bm{Z}_{<k} \cup \{\bm{z}_k\}$
        
        \STATE \textcolor{blue}{\textit{// 3. Compute Step-wise Losses}} \textcolor{red}{\textit{(corresponds to the last two terms in~\eqref{eq: learning})}}
        \STATE \textbf{Latent Reasoning Loss:} Sample a token $r_{j} \in \mathcal{R}_k$ and compute $\mathcal{L}_{reasoning}^{(k)} \leftarrow -\log p_{\psi}(r_{j} \mid \bm{z}_k)$
        \STATE \textbf{Regularizer Loss (KL):} Compute $\mathcal{L}_{KL}^{(k)}$ between $\mathcal{N}(\bm{\mu}_k, \bm{\sigma}_k^2)$ and $\mathcal{N}(\hat{\bm{z}}_k, \bm{I})$ using~\eqref{eq:kld2}
        \STATE Accumulate: $\mathcal{L}_{total} \leftarrow \mathcal{L}_{total} + \mathcal{L}_{reasoning}^{(k)} + \mathcal{L}_{KL}^{(k)} $
    \ENDFOR
    
    \STATE \textcolor{blue}{\textit{// 4. Compute Answer Loss}} \textcolor{red}{\textit{(corresponds to the first term in~\eqref{eq: learning})}}
    \STATE Compute $\mathcal{L}_{answer} \leftarrow -\sum_{i=1}^{L_a} \log p_{\theta'}(a_i \mid \mathcal{Q}, \bm{Z}, \mathcal{A}_{<i})$
    \STATE $\mathcal{L}_{total} \leftarrow \mathcal{L}_{total} + \mathcal{L}_{answer}$
    
    \STATE \textbf{Update Parameters:} $\theta', \gamma \leftarrow \text{Optimizer}(\nabla \mathcal{L}_{total})$
\ENDWHILE
\end{algorithmic}
\end{algorithm}

\begin{algorithm}[t]
\caption{Inference Scheme of ReGuLaR}
\label{alg: inference}
\begin{algorithmic}[1]
\REQUIRE Question $\mathcal{Q}$, Trained LLM with parameters $\theta'$=\{$\phi$, $\psi$\}, Max reasoning steps $K_{max}$.
\STATE \textbf{Initialization:} Latent reasoning state history  $\bm{Z}_{<1} \leftarrow \emptyset$, Latent reasoning step $k \leftarrow 1$.

\STATE \textcolor{blue}{\textit{// Phase 1: Latent Reasoning}}
\WHILE{$k \le K_{max}$}
    \STATE Predict posterior parameters: $\bm{\mu}_k, \log\bm{\sigma}_k \leftarrow p_{\phi}(\mathcal{Q}, \bm{Z}_{<k})$
    \STATE Sample latent reasoning state: $\bm{\epsilon} \sim \mathcal{N}(\bm{0}, \bm{I}), \bm{z}_k \leftarrow \bm{\mu}_k + \bm{\sigma}_k \odot \bm{\epsilon}$
    \STATE Decode representative token: $\hat{r} \sim p_{\psi}(r \mid \bm{z}_k)$ 
    \IF{$\hat{r} == \text{\texttt{<EOS\_Reasoning>}}$}
        \STATE \textbf{Break}
    \ENDIF
    \STATE Update history: $\bm{Z}_{<k+1} \leftarrow \bm{Z}_{<k} \cup \{\bm{z}_k\}$
    \STATE $k \leftarrow k + 1$
\ENDWHILE

\STATE \textcolor{blue}{\textit{// Phase 2: Answer Generation}}
\STATE Initialize answer sequence $\mathcal{A}_{<1} \leftarrow \emptyset$, Answer generation step $i \leftarrow 1$.
\WHILE{answer not finished}
    \STATE Sample token: $a_i \sim p_{\theta'}(a \mid \mathcal{Q}, \bm{Z}, \mathcal{A}_{<i})$
    \IF{$a_i == \text{\texttt{<EOS>}}$}
        \STATE \textbf{Break}
    \ENDIF
    \STATE Update answer: $\mathcal{A}_{<i+1} \leftarrow \mathcal{A}_{<i} \cup \{a_i\}$
    \STATE $i \leftarrow i + 1$
\ENDWHILE

\STATE \textbf{return} Answer $\mathcal{A}$
\end{algorithmic}
\end{algorithm}

\textbf{Hyperparameters.}
For the LLM backbone, we primarily leverage LLaMA-3.2-1B-Instruct~\citep{grattafiori2024llama} unless otherwise specified.
Specifically, we keep the LLM backbone frozen and exclusively optimize LoRA~\citep{hulora} modules, which are configured with $r=128$ and $\alpha=32$ following established baselines.
In addition, both the adapter and the latent reasoning head are instantiated as Multi-Layer Perceptrons (MLPs), where the adapter maps the visual encoder's output dimension ($d_v$=1280) to the LLM's hidden dimension ($d_h$=2048) and the latent reasoning head directly operates within the LLM's hidden dimension ($d_h$=2048).
For training, we optimize the model using the AdamW optimizer~\citep{loshchilov2018decoupled} with a weight decay of $0.01$ and a learning rate of 1e-4, employing a constant schedule with a 1,000-step warmup phase.
Specifically, we utilize Distributed Data Parallel across eight NVIDIA A100 GPUs to ensure training stability and efficiency.
For inference, we employ a stochastic generation strategy using nucleus sampling with top-p of 0.9 and temperature of 1.0 to extract answers.
Specifically, we perform five independent runs using distinct random seeds (from 0 to 4) to ensure reproducibility and reliability.

Additionally, the training and inference schemes of ReGuLaR are presented in Algorithms~\ref{alg: regular} and~\ref{alg: inference}.

\begin{table*}[h]
\centering
\caption{Performance comparison of our proposed ReGuLaR across different font sizes, where we report the averaged number and 95\% confidence interval ($\pm$) on Accuracy (Acc. \%) and Reasoning Length (\# L).}
\label{tab:font-size-results}
\setlength{\tabcolsep}{3.95pt}
\small{
\begin{tabular}{ccccccccccc}
\toprule
\multirow{2}{*}[-0.55ex]{\begin{tabular}{@{}l@{}}Font Size\end{tabular}} & \multicolumn{2}{c}{GSM8K-Aug} & \multicolumn{2}{c}{GSM-Hard} & \multicolumn{2}{c}{SVAMP} & \multicolumn{2}{c}{MultiArith} & \multicolumn{2}{c}{Average}\\
\cmidrule(lr){2-3} \cmidrule(lr){4-5} \cmidrule(lr){6-7} \cmidrule(lr){8-9} \cmidrule(lr){10-11}
& \multicolumn{1}{c}{Acc.} & \multicolumn{1}{c}{\#~L} & \multicolumn{1}{c}{Acc.} & \multicolumn{1}{c}{\#~L} & \multicolumn{1}{c}{Acc.} & \multicolumn{1}{c}{\#~L} & \multicolumn{1}{c}{Acc.} & \multicolumn{1}{c}{\#~L} & \multicolumn{1}{c}{Acc.} & \multicolumn{1}{c}{\#~L} \\
\midrule
20pt & $34.4_{\pm0.25}$ & $4.46_{\pm0.17}$ & $8.02_{\pm0.15}$ & $4.11_{\pm0.45}$ & $48.4_{\pm0.23}$ & $2.17_{\pm0.36}$ & $88.4_{\pm0.11}$ & $2.62_{\pm0.28}$ & 44.8 & 3.34 \\
16pt & $33.2_{\pm0.22}$ & $4.15_{\pm0.20}$ & $8.74_{\pm0.07}$ & $5.23_{\pm0.33}$ & $48.9_{\pm0.26}$ & $1.86_{\pm0.15}$ & $87.2_{\pm0.25}$ & $2.09_{\pm0.09}$ & 44.5 & 3.33 \\
12pt & $34.0_{\pm0.33}$ & $4.01_{\pm0.25}$ & $8.51_{\pm0.09}$ & $4.81_{\pm0.56}$ & $51.1_{\pm0.41}$ & $2.51_{\pm0.74}$ & $87.6_{\pm0.19}$ & $2.16_{\pm0.06}$ & 45.3 & 3.37 \\
\rowcolor{highlightred}
9pt & $34.9_{\pm0.26}$ & $3.69_{\pm0.21}$ & $8.27_{\pm0.14}$ & $4.12_{\pm0.48}$ & $50.1_{\pm0.39}$ & $2.02_{\pm0.18}$ & $89.2_{\pm0.27}$ & $2.28_{\pm0.27}$ & 45.6 & 3.03 \\
\bottomrule
\end{tabular}
}
\end{table*}

\section{More Experimental Results.}\label{app: results}
\subsection{Ablation Studies on Rendering Configuration}
In our standard implementation, we adopt the optimal rendering configuration (i.e., those summarized in Table~\ref{tab:rendering_config}) identified in Glyph~\citep{cheng2025glyph} to ensure consistent visual encoding and maximize the semantic density.
Here, to verify the robustness and generalizability of our method across varying rendering configurations, we conduct ablation studies on two pivotal rendering parameters: 

\textbf{Font Size.}
We compare the performance of our proposed ReGuLaR by varying the font size from 9pt to 20pt.
As presented in Table~\ref{tab:font-size-results}, ReGuLaR demonstrates remarkable stability across different font sizes. 
Specifically, despite substantial variations in font size, the average accuracy fluctuates only marginally (ranging from 44.5\% to 45.6\%), while the reasoning length remains largely consistent.
These results indicate that our method effectively captures semantic information regardless of the text's scale, ensuring robust performance without requiring precise font size tuning.

\textbf{Rendering Density (DPI).}
We investigate the sensitivity of our proposed ReGuLaR to information density by adjusting the DPI from 72 to 300.
As presented in Table~\ref{tab:dpi-results}, the performance remains highly consistent across this wide range. 
Notably, ReGuLaR achieves comparable average accuracy at both the lowest density (45.6\% at 72 DPI) and the highest density (45.2\% at 300 DPI). 
This suggests that ReGuLaR is resilient to variations in image clarity and pixel density, capable of extracting reliable features under diverse DPI settings.

\begin{table*}[t]
\centering
\caption{Performance comparison of our proposed ReGuLaR across different rendering density (DPI) settings, where we report the averaged number and 95\% confidence interval ($\pm$) on Accuracy (Acc. \%) and Reasoning Length (\# L).}
\label{tab:dpi-results}
\setlength{\tabcolsep}{4.9pt}
\small{
\begin{tabular}{ccccccccccc}
\toprule
\multirow{2}{*}[-0.55ex]{DPI} & \multicolumn{2}{c}{GSM8K-Aug} & \multicolumn{2}{c}{GSM-Hard} & \multicolumn{2}{c}{SVAMP} & \multicolumn{2}{c}{MultiArith} & \multicolumn{2}{c}{Average}\\
\cmidrule(lr){2-3} \cmidrule(lr){4-5} \cmidrule(lr){6-7} \cmidrule(lr){8-9} \cmidrule(lr){10-11}
& \multicolumn{1}{c}{Acc.} & \multicolumn{1}{c}{\#~L} & \multicolumn{1}{c}{Acc.} & \multicolumn{1}{c}{\#~L} & \multicolumn{1}{c}{Acc.} & \multicolumn{1}{c}{\#~L} & \multicolumn{1}{c}{Acc.} & \multicolumn{1}{c}{\#~L} & \multicolumn{1}{c}{Acc.} & \multicolumn{1}{c}{\#~L} \\
\midrule
300 & $33.3_{\pm0.18}$ & $3.93_{\pm0.35}$ & $8.07_{\pm0.15}$ & $3.73_{\pm0.41}$ & $50.7_{\pm0.20}$ & $2.16_{\pm0.28}$ & $88.9_{\pm0.28}$ & $2.24_{\pm0.31}$ & 45.2 & 3.02 \\
144 & $33.4_{\pm0.07}$ & $4.87_{\pm0.36}$ & $7.67_{\pm0.06}$ & $5.05_{\pm0.39}$ & $49.9_{\pm0.42}$ & $2.19_{\pm0.31}$ & $87.5_{\pm0.12}$ & $3.07_{\pm0.19}$ & 44.6 & 3.80 \\
96 & $32.6_{\pm0.16}$ & $4.13_{\pm0.16}$ & $7.87_{\pm0.13}$ & $5.19_{\pm0.42}$ & $48.2_{\pm0.11}$ & $2.27_{\pm0.03}$ & $88.3_{\pm0.16}$ & $2.12_{\pm0.04}$ & 44.2 & 3.43 \\
\rowcolor{highlightred}
72 & $34.9_{\pm0.26}$ & $3.69_{\pm0.21}$ & $8.27_{\pm0.14}$ & $4.12_{\pm0.48}$ & $50.1_{\pm0.39}$ & $2.02_{\pm0.18}$ & $89.2_{\pm0.27}$ & $2.28_{\pm0.27}$ & 45.6 & 3.03 \\
\bottomrule
\end{tabular}
}
\end{table*}

\subsection{Ablation Studies on Visual Encoder Modes}
In our standard implementation, we employ the trained visual encoder from DeepSeek-OCR~\citep{wei2025deepseek_OCR} to encode each rendered image into a sequence of visual tokens, which are subsequently aggregated via mean pooling to derive a single and compact visual-semantic representation.
In particular, this trained visual encoder supports four modes: Tiny, Small, Base, and Large, corresponding to resolutions of $512\times512$, $640\times640$, $1024\times1024$, and $1280\times1280$, resulting in 64, 100, 256, and 400 vision tokens.
Depending on the selected mode, the input rendered images are processed via either adaptive resizing (for Tiny/Small) or padding (for Base/Large) to align with the specific resolution.
Therefore, while the input rendered images remain identical in their original resolution, selecting different modes forces the visual encoder to process them at varying internal resolutions and token budgets.
Here, we conduct ablation studies across these four modes to investigate the impact of visual encoding granularity on our method.

As presented in Table~\ref{tab:encoder-size-results}, the results reveal a counter-intuitive yet profound insight into the visual-semantic compression mechanism: the Tiny mode, despite internally resizing the input rendered images to $512\times512$ and utilizing only 64 intermediate tokens, achieves performance comparable to that of the high-resolution modes.
We attribute this remarkable robustness to the information aggregation nature of our method. 
Since the intermediate visual tokens are ultimately pooled into a single and compact visual-semantic representation, it relies more on high-level semantic abstraction rather than fine-grained pixel-level details.
Therefore, we adopt the Tiny mode as the default configuration in our experiments, effectively reducing the visual processing overhead by approximately $6 \times$ compared with the Large mode while ensuring that the final visual representation remains semantically rich and accurate.

\begin{table*}[t]
\centering
\caption{Performance comparison of our proposed ReGuLaR across different visual encoder modes, where we report the averaged number and 95\% confidence interval ($\pm$) on Accuracy (Acc. \%) and Reasoning Length (\# L).}
\label{tab:encoder-size-results}
\setlength{\tabcolsep}{4.55pt}
\small{
\begin{tabular}{ccccccccccc}
\toprule
\multirow{2}{*}[-0.55ex]{Mode} & \multicolumn{2}{c}{GSM8K-Aug} & \multicolumn{2}{c}{GSM-Hard} & \multicolumn{2}{c}{SVAMP} & \multicolumn{2}{c}{MultiArith} & \multicolumn{2}{c}{Average}\\
\cmidrule(lr){2-3} \cmidrule(lr){4-5} \cmidrule(lr){6-7} \cmidrule(lr){8-9} \cmidrule(lr){10-11}
& \multicolumn{1}{c}{Acc.} & \multicolumn{1}{c}{\#~L} & \multicolumn{1}{c}{Acc.} & \multicolumn{1}{c}{\#~L} & \multicolumn{1}{c}{Acc.} & \multicolumn{1}{c}{\#~L} & \multicolumn{1}{c}{Acc.} & \multicolumn{1}{c}{\#~L} & \multicolumn{1}{c}{Acc.} & \multicolumn{1}{c}{\#~L} \\
\midrule
Large & $33.0_{\pm0.36}$ & $3.77_{\pm0.24}$ & $7.35_{\pm0.17}$ & $4.28_{\pm0.29}$ & $48.5_{\pm0.41}$ & $2.42_{\pm0.25}$ & $88.4_{\pm0.38}$ & $3.08_{\pm0.12}$ & 44.3 & 3.39 \\
Base & $34.3_{\pm0.46}$ & $3.86_{\pm0.48}$ & $7.67_{\pm0.09}$ & $4.60_{\pm0.36}$ & $51.6_{\pm0.33}$ & $2.23_{\pm0.22}$ & $88.6_{\pm0.41}$ & $2.41_{\pm0.32}$ & 45.5 & 3.28 \\
Small & $34.0_{\pm0.29}$ & $3.64_{\pm0.34}$ & $7.53_{\pm0.12}$ & $4.07_{\pm0.45}$ & $52.1_{\pm0.31}$ & $2.71_{\pm0.29}$ & $89.5_{\pm0.23}$ & $2.23_{\pm0.28}$ & 45.8 & 3.16 \\
\rowcolor{highlightred}
Tiny & $34.9_{\pm0.26}$ & $3.69_{\pm0.21}$ & $8.27_{\pm0.14}$ & $4.12_{\pm0.48}$ & $50.1_{\pm0.39}$ & $2.02_{\pm0.18}$ & $89.2_{\pm0.27}$ & $2.28_{\pm0.27}$ & 45.6 & 3.03 \\
\bottomrule
\end{tabular}
}
\end{table*}

\subsection{Ablation Studies on Learning Paradigms}
In our standard implementation, we train the proposed ReGuLaR via the unified objective function defined in~\eqref{eq: learning}, which integrates three critical components to optimize the model jointly: the answer generation loss ($\mathcal{L}_\text{answer}^{\text{Latent}}$) directly ensures answer correctness, the latent reasoning loss ($\mathcal{L}_\text{reasoning}^{\text{Latent}}$) preserves semantic integrity, and the KL divergence term ($\mathcal{L}_\text{KL}^{\text{Latent}}$) regularizes the posterior distribution.
To investigate the distinct contribution of each component, we conduct ablation studies by selectively removing the latter two terms. 
Notably, the answer generation loss (i.e., the first term in~\eqref{eq: learning}) is retained across all variants, as it serves as the fundamental supervision signal for the reasoning task.

As presented in Table~\ref{tab:learning-results}, the absence of the KL divergence term ($\mathcal{L}_\text{KL}^{\text{Latent}}$) leads to catastrophic failure (i.e., accuracy$<14\%$), regardless of the latent reasoning loss.
In stark contrast, introducing $\mathcal{L}_\text{KL}^{\text{Latent}}$ alone significantly boosts performance to 41.9\%.
This result empirically corroborates our analysis in Section~\ref{ssec: vae}: without the constraint imposed by the prior distribution, neither the distant supervision from the final answer nor the semantic supervision from textual reconstruction is sufficient.
In addition, combining all components achieves the peak performance of 45.6\%, demonstrating that the semantic richness from text reconstruction and the structural guidance from distribution regularization are synergistic and mutually indispensable.

\begin{table*}[t]
\centering
\caption{Performance comparison of our proposed ReGuLaR across different learning paradigms, where we report the averaged number and 95\% confidence interval ($\pm$) on Accuracy (Acc. \%) and Reasoning Length (\# L).}
\label{tab:learning-results}
\setlength{\tabcolsep}{2.7pt}
\small{
\begin{tabular}{cccccccccccc}
\toprule
\multirow{2}{*}[-0.55ex]{$\mathcal{L}_\text{KL}^{\text{Latent}}$} & \multirow{2}{*}[-0.55ex]{$\mathcal{L}_\text{reasoning}^{\text{Latent}}$} & \multicolumn{2}{c}{GSM8K-Aug} & \multicolumn{2}{c}{GSM-Hard} & \multicolumn{2}{c}{SVAMP} & \multicolumn{2}{c}{MultiArith} & \multicolumn{2}{c}{Average}\\
\cmidrule(lr){3-4} \cmidrule(lr){5-6} \cmidrule(lr){7-8} \cmidrule(lr){9-10} \cmidrule(lr){11-12}
& & \multicolumn{1}{c}{Acc.} & \multicolumn{1}{c}{\#~L} & \multicolumn{1}{c}{Acc.} & \multicolumn{1}{c}{\#~L} & \multicolumn{1}{c}{Acc.} & \multicolumn{1}{c}{\#~L} & \multicolumn{1}{c}{Acc.} & \multicolumn{1}{c}{\#~L} & \multicolumn{1}{c}{Acc.} & \multicolumn{1}{c}{\#~L} \\
\midrule
$\times$ & $\times$ & $5.69_{\pm0.13}$ & $4.73_{\pm0.03}$ & $1.46_{\pm0.11}$ & $5.11_{\pm0.05}$ & $35.7_{\pm1.22}$ & $2.94_{\pm0.03}$ & $8.33_{\pm0.11}$ & $2.40_{\pm0.23}$ & 12.8 & 3.81 \\
$\times$ & $\checkmark$ & $6.52_{\pm0.26}$ & $4.57_{\pm0.16}$ & $1.68_{\pm0.16}$ & $5.54_{\pm0.12}$ & $34.4_{\pm0.81}$ & $1.86_{\pm0.05}$ & $9.93_{\pm0.39}$ & $1.93_{\pm0.19}$ & 13.1 & 3.47\\
$\checkmark$ & $\times$ & $27.3_{\pm0.38}$ & $3.87_{\pm0.10}$ & $6.03_{\pm0.22}$ & $4.71_{\pm0.34}$ & $47.7_{\pm0.27}$ & $1.84_{\pm0.06}$ & $86.4_{\pm0.57}$ & $1.96_{\pm0.14}$ & 41.9 & 3.10 \\
\rowcolor{highlightred}
$\checkmark$ & $\checkmark$ & $34.9_{\pm0.26}$ & $3.69_{\pm0.21}$ & $8.27_{\pm0.14}$ & $4.12_{\pm0.48}$ & $50.1_{\pm0.39}$ & $2.02_{\pm0.18}$ & $89.2_{\pm0.27}$ & $2.28_{\pm0.27}$ & 45.6 & 3.03 \\
\bottomrule
\end{tabular}
}
\end{table*}

\subsection{Ablation Studies on Modeling Strategies}
In our standard implementation, we model the latent reasoning process as a probabilistic transition (i.e., expressed in~\eqref{eq:posterior}), leveraging the latent reasoning head to predict the distribution parameters $\bm{\mu}$ and $\log\bm{\sigma}$ of the next latent reasoning state.
To verify the effectiveness of this probabilistic modeling strategy, we conduct ablation studies comparing it against the deterministic variant.
Specifically, this deterministic variant directly leverages the latent reasoning head to predict the next latent reasoning state, which is functionally equivalent to a greedy strategy that always selects the most likely mean vector.

As presented in Table~\ref{tab:modeling-results}, the results demonstrate the clear superiority of the probabilistic modeling strategy. 
Specifically, it outperforms the deterministic variant across all datasets, achieving an average accuracy of 45.6\%. 
We attribute the performance drop in the deterministic variant to the ``mean collapse" phenomenon. 
Since the reasoning process often allows for multiple valid subsequent steps, a deterministic predictor tends to output the average of all possible outcomes to minimize the reconstruction error. This results in blurred semantic representations that fail to capture the precise logic required for complex reasoning. In contrast, our probabilistic strategy models the underlying distribution, enabling the sampling of sharp and distinct latent reasoning states that preserve semantic integrity.

\begin{table*}[t]
\centering
\caption{Performance comparison of our proposed ReGuLaR across different modeling strategies, where we report the averaged number and 95\% confidence interval ($\pm$) on Accuracy (Acc. \%) and Reasoning Length (\# L).}
\label{tab:modeling-results}
\setlength{\tabcolsep}{3.2pt}
\small{
\begin{tabular}{ccccccccccc}
\toprule
\multirow{2}{*}[-0.55ex]{Strategy} & \multicolumn{2}{c}{GSM8K-Aug} & \multicolumn{2}{c}{GSM-Hard} & \multicolumn{2}{c}{SVAMP} & \multicolumn{2}{c}{MultiArith} & \multicolumn{2}{c}{Average}\\
\cmidrule(lr){2-3} \cmidrule(lr){4-5} \cmidrule(lr){6-7} \cmidrule(lr){8-9} \cmidrule(lr){10-11}
& \multicolumn{1}{c}{Acc.} & \multicolumn{1}{c}{\#~L} & \multicolumn{1}{c}{Acc.} & \multicolumn{1}{c}{\#~L} & \multicolumn{1}{c}{Acc.} & \multicolumn{1}{c}{\#~L} & \multicolumn{1}{c}{Acc.} & \multicolumn{1}{c}{\#~L} & \multicolumn{1}{c}{Acc.} & \multicolumn{1}{c}{\#~L} \\
\midrule
Deterministic & $32.6_{\pm0.34}$ & $3.45_{\pm0.02}$ & $7.24_{\pm0.06}$ & $5.65_{\pm0.52}$ & $48.9_{\pm0.37}$ & $2.72_{\pm0.25}$ & $88.1_{\pm0.14}$ & $2.22_{\pm0.29}$ & 44.2 & 3.51 \\
\rowcolor{highlightred}
Probabilistic & $34.9_{\pm0.26}$ & $3.69_{\pm0.21}$ & $8.27_{\pm0.14}$ & $4.12_{\pm0.48}$ & $50.1_{\pm0.39}$ & $2.02_{\pm0.18}$ & $89.2_{\pm0.27}$ & $2.28_{\pm0.27}$ & 45.6 & 3.03 \\
\bottomrule
\end{tabular}
}
\end{table*}

\subsection{Ablation Studies on Regularization Strategies}
In our standard implementation, we decompose the reasoning chain into $K$ segments and subsequently render them into images to yield $K$ visual representations, which serve as dense semantic anchors to regularize the posterior distribution of the latent reasoning state during training (i.e., illustrated in Figure~\ref{fig: Concept}).
To verify the distinct contribution of this vision-based regularization, we conduct ablation studies comparing it against the text-based variant.
Specifically, this text-based variant directly aggregates the embeddings of tokens within the same segment into one textual representation to regularize the posterior distribution of the latent reasoning state, while keeping all other settings invariant.

As presented in Table~\ref{tab:regularing-results}, the results demonstrate a substantial performance advantage for the vision-based regularization strategy. 
Specifically, it significantly outperforms the text-based variant across all datasets, increasing the average accuracy from 42.3\% to 45.6\%.
We attribute this superiority to the dense information compression capability of the visual modality.
The text-based variant, which relies on pooling token embeddings, tends to dilute the structural and topological details of the reasoning chain (e.g., the spatial layout of arithmetic operations), resulting in a ``blurred" semantic target. 
In contrast, the vision-based approach compels the model to align its latent reasoning state with the corresponding rendered image, which serves as a highly compact and structured semantic anchor. 
This cross-modal constraint forces the model to capture the holistic logic of the segment rather than just the average meaning of its tokens, thereby providing a more robust signal for regularization.
Notably, as discussed in Section~\ref{ssec: advantage}, these visual representations are strictly confined to the training phase, meaning that they incur no additional cost during inference.

\begin{table*}[t]
\centering
\caption{Performance comparison of our proposed ReGuLaR across different regularization strategies, where we report the averaged number and 95\% confidence interval ($\pm$) on Accuracy (Acc. \%) and Reasoning Length (\# L).}
\label{tab:regularing-results}
\setlength{\tabcolsep}{3.3pt}
\small{
\begin{tabular}{ccccccccccc}
\toprule
\multirow{2}{*}[-0.55ex]{Strategy} & \multicolumn{2}{c}{GSM8K-Aug} & \multicolumn{2}{c}{GSM-Hard} & \multicolumn{2}{c}{SVAMP} & \multicolumn{2}{c}{MultiArith} & \multicolumn{2}{c}{Average}\\
\cmidrule(lr){2-3} \cmidrule(lr){4-5} \cmidrule(lr){6-7} \cmidrule(lr){8-9} \cmidrule(lr){10-11}
& \multicolumn{1}{c}{Acc.} & \multicolumn{1}{c}{\#~L} & \multicolumn{1}{c}{Acc.} & \multicolumn{1}{c}{\#~L} & \multicolumn{1}{c}{Acc.} & \multicolumn{1}{c}{\#~L} & \multicolumn{1}{c}{Acc.} & \multicolumn{1}{c}{\#~L} & \multicolumn{1}{c}{Acc.} & \multicolumn{1}{c}{\#~L} \\
\midrule
Text-based & $28.3_{\pm0.20}$ & $3.53_{\pm0.31}$ & $6.53_{\pm0.11}$ & $4.21_{\pm0.03}$ & $47.5_{\pm0.18}$ & $1.97_{\pm0.03}$ & $86.9_{\pm0.37}$ & $2.07_{\pm0.13}$ & 42.3 & 2.95 \\
\rowcolor{highlightred}
Vision-based & $34.9_{\pm0.26}$ & $3.69_{\pm0.21}$ & $8.27_{\pm0.14}$ & $4.12_{\pm0.48}$ & $50.1_{\pm0.39}$ & $2.02_{\pm0.18}$ & $89.2_{\pm0.27}$ & $2.28_{\pm0.27}$ & 45.6 & 3.03 \\
\bottomrule
\end{tabular}
}
\end{table*}

\begin{figure*}[t]
    \centering 
    \includegraphics[width=\textwidth]{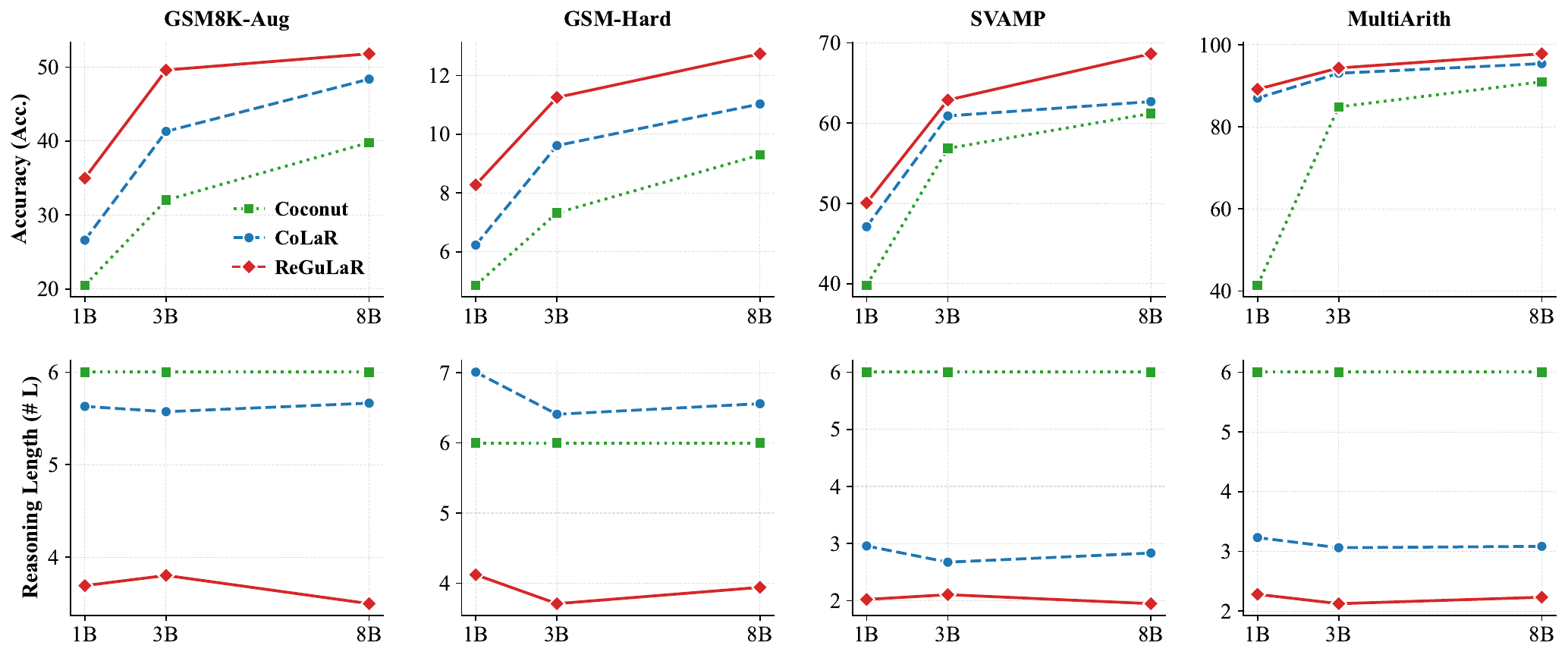} 
    \caption{Scalability analysis across varying model sizes, where we employ LLaMA-3.2 (1B, 3B) and LLaMA-3.1 (8B) Instruct variants as the LLM backbones.}
    \label{fig: Full_Scalability}
\end{figure*}

\subsection{Further Scalability Analysis of ReGuLaR}
Figure~\ref{fig: Full_Scalability} further illustrates the performance of ReGuLaR across varying model sizes, where ReGuLaR demonstrates strong positive scaling behavior. 
Specifically, compared with the top-performing baselines (i.e., CoLaR and Coconut), ReGuLaR consistently maintains the highest accuracy with the shortest reasoning length across all model scales and datasets, highlighting its seamless scalability and potential for broader application in large-scale foundation models.

\section{Examples of Rendered Reasoning Chains}
\begin{figure}[H]
\centering
\begin{subfigure}[t]{0.48\textwidth}
\centering
\includegraphics[width=\linewidth]{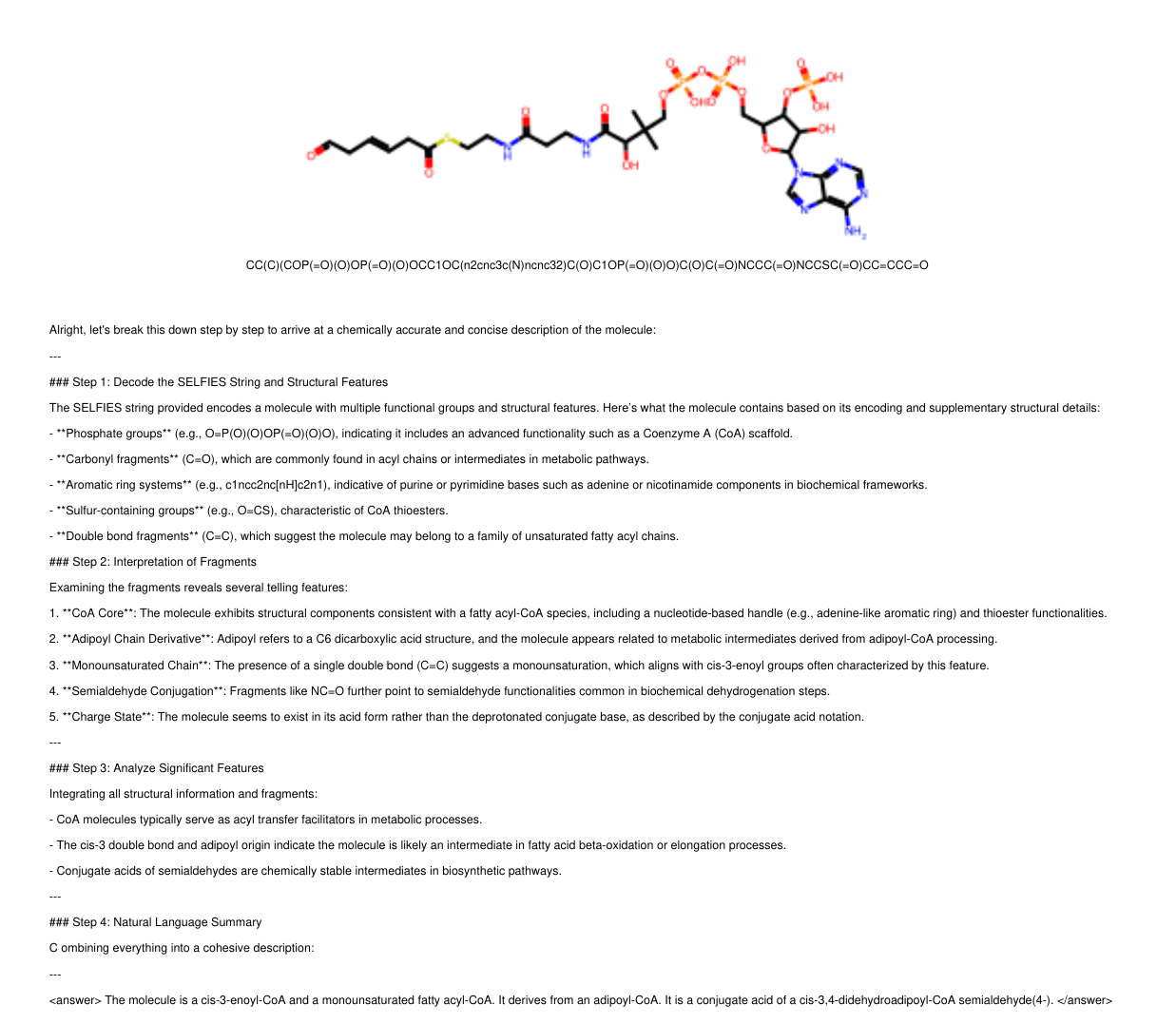}
\caption{Rendered reasoning chains with 2D (example 1)}
\label{fig:example_molecular_a}
\end{subfigure}
\hfill
\begin{subfigure}[t]{0.48\textwidth}
\centering
\includegraphics[width=\linewidth]{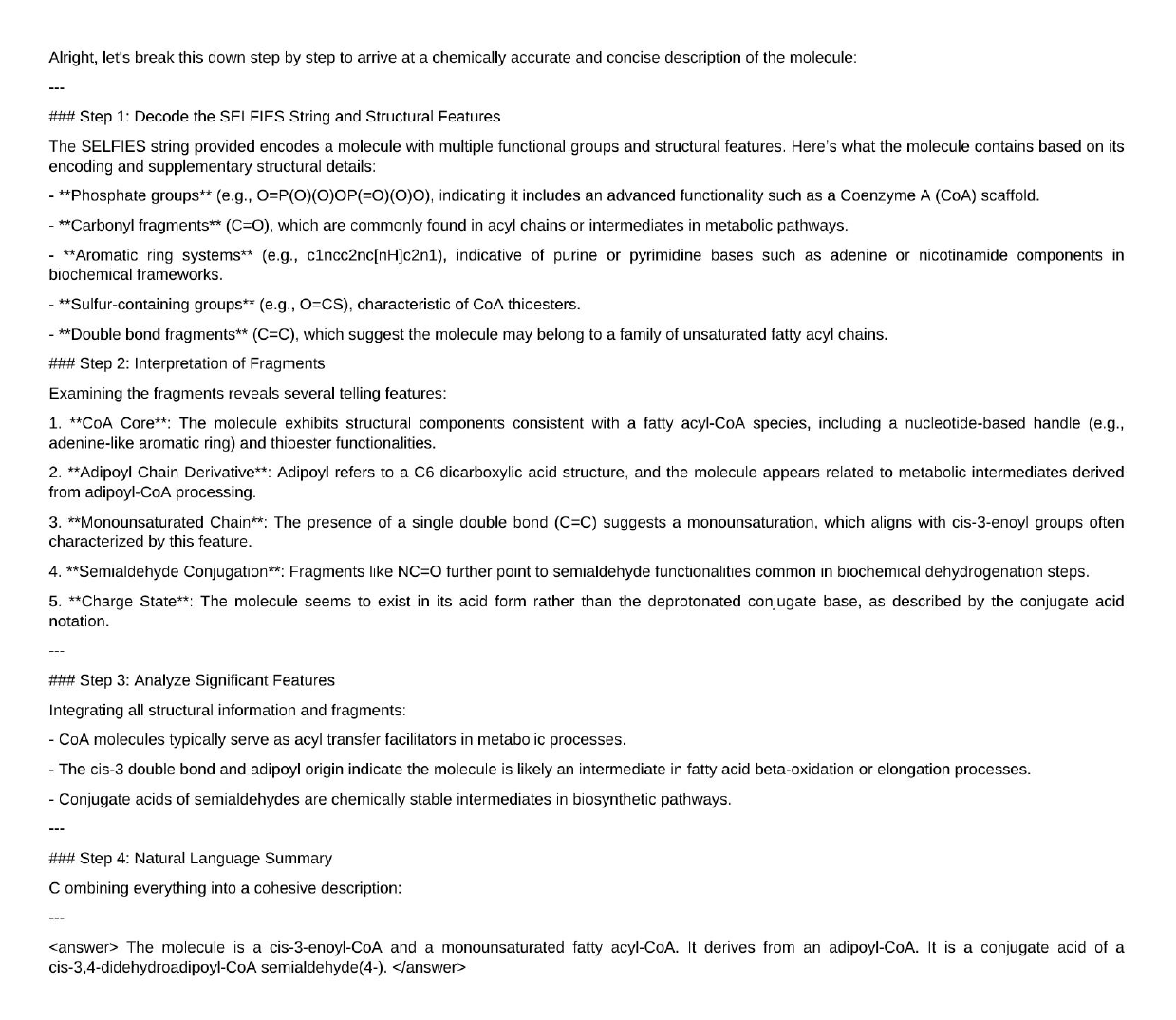}
\caption{Rendered reasoning chains without 2D (example 1)}
\label{fig:example_molecular_b}
\end{subfigure}

\begin{subfigure}[t]{0.48\textwidth}
    \centering
    \includegraphics[width=\linewidth]{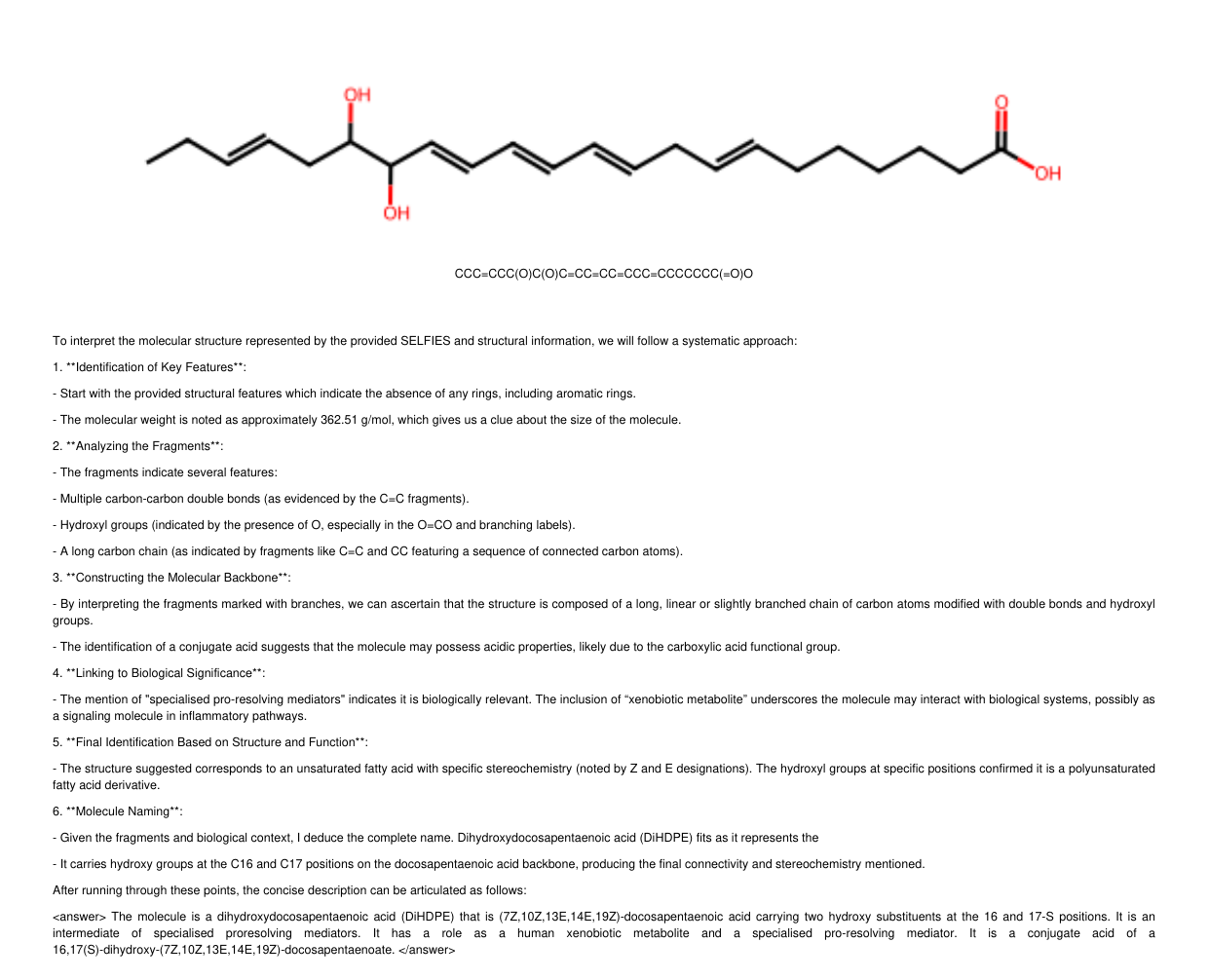}
    \caption{Rendered reasoning chains with 2D (example 2)}
    \label{fig:example_molecular_c}
\end{subfigure}
\hfill
\begin{subfigure}[t]{0.48\textwidth}
    \centering
    \includegraphics[width=\linewidth]{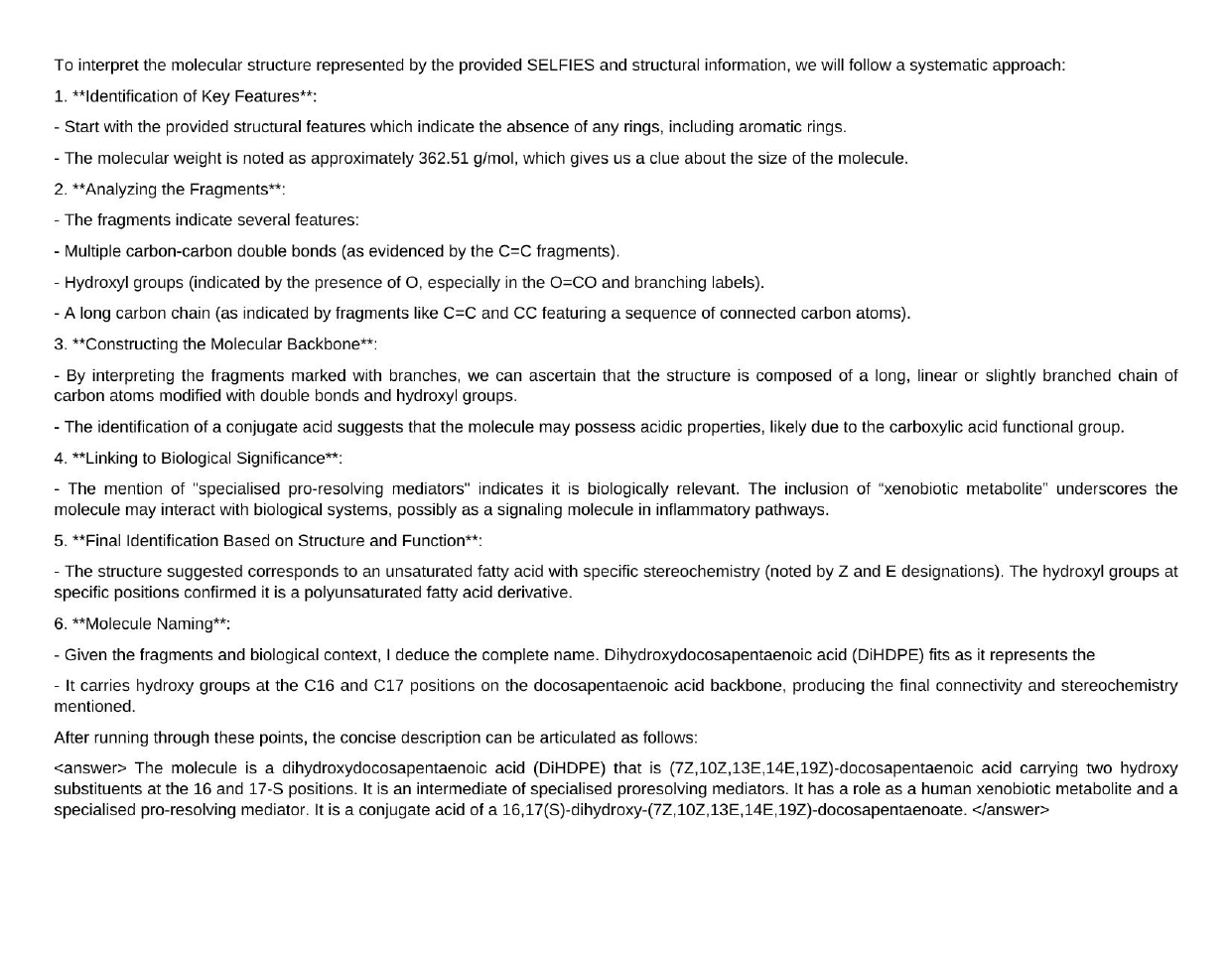}
    \caption{Rendered reasoning chains without 2D (example 2)}
    \label{fig:example_molecular_d}
\end{subfigure}

\caption{Examples of rendered reasoning chains for the molecular captioning task. 
}
\label{fig:example_molecular_all}
\end{figure}

In the molecular captioning task, we utilize RDKit to generate the corresponding 2D molecular graph for each textual reasoning chain, thereby constructing multi-modal reasoning chains.
Subsequently, these reasoning chains are rendered into images, from which visual representations are extracted to regularize the posterior distribution of the latent reasoning state during training.
Here, Figure~\ref{fig:example_molecular_all} presents two examples of rendered reasoning chains, each shown in two variants: rendered multi-model reasoning chains with explicit 2D molecular graphs and original textual reasoning chains without 2D molecular graphs. 
For rendered multi-model reasoning chains that include 2D molecular graphs (e.g., Figures~\ref{fig:example_molecular_a} and~\ref{fig:example_molecular_c}), we position the corresponding 2D molecular graph at the top of the rendered image and annotate it below with its SMILES string. 
The remainder of the rendering is identical to the counterpart without 2D graphs, consisting solely of textual reasoning steps.

\end{document}